\documentclass[10pt,twocolumn,letterpaper]{article}

\usepackage{cvpr} 
   
\usepackage[dvipsnames]{xcolor}

\definecolor{cvprblue}{rgb}{0.21,0.49,0.74}
\usepackage[pagebackref,breaklinks,colorlinks,citecolor=cvprblue]{hyperref}
\newcommand{\myPara}[1]{\vspace{.05in}\noindent\textbf{#1}}
\usepackage{tikz}
\usepackage{bm}
\usepackage{bbm}
\usepackage{multirow}
\usepackage{xcolor}
\usepackage{tablefootnote}
\usepackage{algorithm}
\usepackage{algpseudocode}
\definecolor{blue}{RGB}{82, 115, 197}
\definecolor{green}{RGB}{133, 187, 84}
\definecolor{red}{RGB}{176, 36, 24}
\def\paperID{902} 
\def\confName{CVPR}
\def\confYear{2025}

\title{LaVin-DiT: Large Vision Diffusion Transformer}

\author{
Zhaoqing Wang\textsuperscript{1,4} \quad
Xiaobo Xia\textsuperscript{2} \quad
Runnan Chen\textsuperscript{1} \quad
Dongdong Yu\textsuperscript{4} \\[0.5em]
Changhu Wang\textsuperscript{4}\thanks{Corresponding author.} \quad
Mingming Gong\textsuperscript{3}\footnotemark[1] \quad
Tongliang Liu\textsuperscript{1}\footnotemark[1] \\[1em]
\small
\textsuperscript{1}The University of Sydney \quad
\textsuperscript{2}National University of Singapore \quad
\textsuperscript{3}The University of Melbourne \quad
\textsuperscript{4}AIsphere \\[0.5em]
{\tt\small wangchanghu@aishi.ai, mingming.gong@unimelb.edu.au, tongliang.liu@sydney.edu.au}\\
{\tt\small\url{https://derrickwang005.github.io/LaVin-DiT/}}
}
\begin{document}
\maketitle
\begin{abstract}
This paper presents the Large Vision Diffusion Transformer (LaVin-DiT), a scalable and unified foundation model designed to tackle over 20 computer vision tasks in a generative framework. Unlike existing large vision models directly adapted from natural language processing architectures, which rely on less efficient autoregressive techniques and disrupt spatial relationships essential for vision data, LaVin-DiT introduces key innovations to optimize generative performance for vision tasks. First, to address the high dimensionality of visual data, we incorporate a spatial-temporal variational autoencoder that encodes data into a continuous latent space. Second, for generative modeling, we develop a joint diffusion transformer that progressively produces vision outputs. Third, for unified multi-task training, in-context learning is implemented. Input-target pairs serve as task context, which guides the diffusion transformer to align outputs with specific tasks within the latent space. During inference, a task-specific context set and test data as queries allow LaVin-DiT to generalize across tasks without fine-tuning. Trained on extensive vision datasets, the model is scaled from 0.1B to 3.4B parameters, demonstrating substantial scalability and state-of-the-art performance across diverse vision tasks. This work introduces a novel pathway for large vision foundation models, underscoring the promising potential of diffusion transformers. The code and models are available.
\end{abstract}    
\section{Introduction}
\label{sec:intro}
\begin{figure}
    \centering
    \includegraphics[width=.99\linewidth]{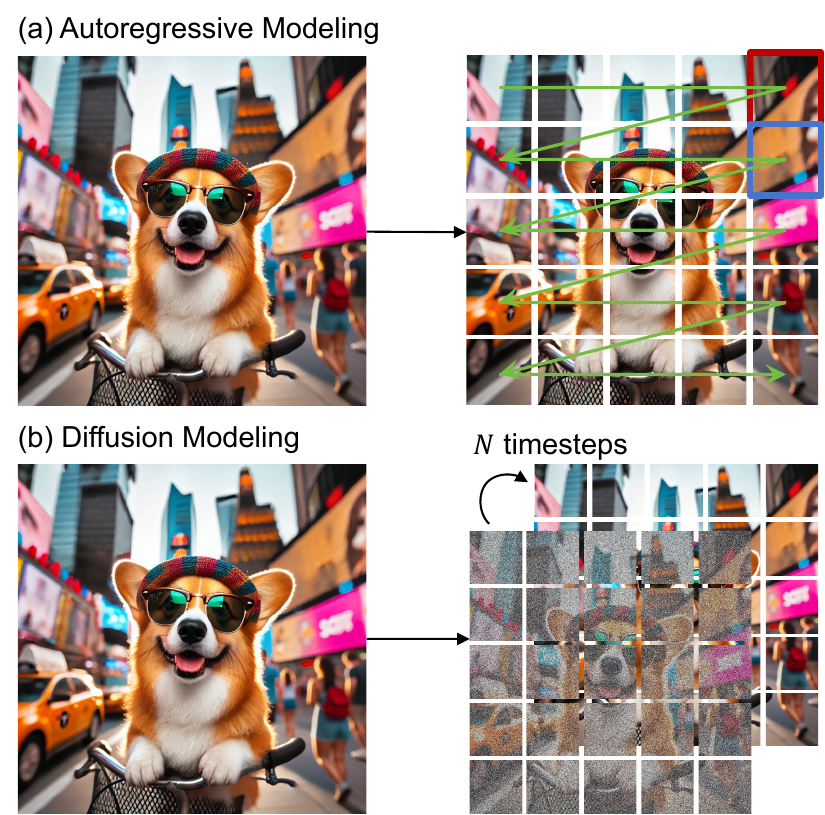}
    \caption{\textbf{Comparison of autoregressive and diffusion modeling.}
    (a) In \textbf{autoregressive modeling}, visual data is divided into a sequence of patches and transformed into a one-dimensional sequence. The model then predicts each token sequentially from left to right and top to bottom, which is computationally intensive for high-dimensional visual data. Besides, tokens marked in \textcolor{red}{red} and \textcolor{blue}{blue} illustrate disrupted spatial dependencies, highlighting the limitations of preserving spatial coherence.
    (b) In contrast, \textbf{diffusion modeling} denoises all tokens in parallel across $N$ timesteps, significantly improving computational efficiency and preserving essential spatial structures crucial for high-performance vision tasks.
    }
    \label{fig:st_vae}
\end{figure}

Large language models (LLMs) like GPT~\cite{brown2020language} and LLaMA~\cite{touvron2023llama} have rapidly gained widespread attention and transformed the field, demonstrating the strong capability to handle a wide range of language tasks within a unified framework~\cite{achiam2023gpt}. This breakthrough of integrating diverse language tasks into a single large model has sparked momentum to develop similar large models for computer vision. The potential to create large vision models~(LVMs) capable of generalizing across multiple vision tasks represents a promising step toward a more versatile, scalable, and efficient approach to vision-based AI~\cite{bar2022visual,lu2022unified,wang2023images,bai2024sequential,luo2024deem}.

However, constructing LVMs presents greater complexity than LLMs due to the inherently diverse and high-dimensional nature of vision data, as well as the need to handle variations in scale, perspective, and lighting across tasks~\cite{ren2016faster,kirillov2019panoptic,zhang2016colorful,silberman2012indoor,wang2024open,wang2022cris}. To handle the problem, recent work~\cite{bai2024sequential} has developed a sequential modeling method that learns from purely vision data by representing images, videos, and annotations in a unified ``visual sentence'' format. This method enables the model to predict sequential vision tokens from a vast dataset, entirely independent of language-based inputs (see Figure~\ref{fig:st_vae}(a)). Although this method has shown promising results in diverse vision tasks, it faces two primary challenges. Specifically, the first issue concerns the efficiency limitations inherent in autoregressive sequence modeling~\cite{sutskever2014sequence}, as it demands token-by-token prediction, which is computationally intensive for high-dimensional vision data~\cite{rombach2022high,wei2024training}. The second issue is the disruption of spatial coherence when converting vision data into a sequential format, which compromises the preservation of spatial dependencies crucial for performance in vision tasks~\cite{zhuo2024lumina}.

In this paper, we introduce a large vision diffusion transformer~(LaVin-DiT) to advance the development of next-generation LVMs. LaVin-DiT enjoys better computational efficiency and effectively preserves spatial relationships within vision data, thereby achieving superior performance across diverse vision tasks~(see Figure~\ref{fig:st_vae}(b)). Technically, to tackle the high-dimensional nature of vision data, we introduce a spatial-temporal variational autoencoder~\cite{kingma2013auto} that encodes data (\ie, image and video) into a continuous latent space, allowing compact representation while preserving essential spatial and temporal features. This reduces computational demands and improves efficiency without sacrificing the model’s ability to capture complex patterns. Besides, for generative modeling, we augment an existing diffusion transformer and propose a joint diffusion transformer with full-sequence joint attention. This module synthesizes visual outputs through parallel denoising steps, effectively reducing sequential dependencies to enhance processing efficiency while maintaining the spatial coherence essential for vision tasks. Moreover, to support unified multi-task training~\cite{vandenhende2021multi}, we incorporate in-context learning~\cite{brown2020language,dong2022survey,weifinetuned,zhangideal,lin2025understanding}, where input-target pairs guide the diffusion transformer in aligning outputs with specific tasks. During inference, LaVin-DiT leverages task-specific context sets and test data as queries to adapt to various tasks without fine-tuning. This capability enables LaVin-DiT to achieve robust generalization across diverse tasks, leading to a versatile solution for complex vision applications.

We conduct comprehensive experiments to demonstrate the superiority of LaVin-DiT. Results show that LaVin-DiT significantly outperforms the strongest baseline LVM~\cite{bai2024sequential} on various vision benchmarks. For instance, it achieves a 24 lower AbsRel in NYU-v2 depth estimation~\cite{silberman2012indoor}. Besides, LaVin-DiT offers $1.7\sim2.3 \times$ faster inference speeds than LVM~\cite{bai2024sequential} across resolutions ranging from $256 \times 256$ to $512 \times 512$. Evaluations across different model sizes showcase the scalability and fast convergence of LaVin-DiT across multiple complex vision tasks. Finally, we observe that increasing the task context length consistently enhances performance across a diverse array of tasks. These promising results establish LaVin-DiT as a highly scalable, efficient, and versatile model, showing a new pathway for large vision foundation models.

\section{Related Work}
\label{sec:related_work}
\myPara{Large vision model.}
Developing a universal framework for diverse tasks across information sources is a longstanding goal in deep learning~\cite{lu2022unified}. Natural language processing has achieved this with ChatGPT\footnote{\url{https://openai.com/index/chatgpt/}} that demonstrates versatility across numerous language tasks, \eg, summarization, reasoning, and translation. In contrast, computer vision is relatively lacking in universal frameworks, largely due to the complexity and diversity of visual data and tasks. Existing methods of universal vision frameworks generally follow two main pathways: image-resembling generation~\cite{chen2021pix2seq,wang2023images,bar2022visual} and sequential modeling~\cite{kolesnikov2022uvim,bai2024sequential}.

The image-resembling generation methods reformulate visual tasks as image generation problems, which allows models to handle dense visual predictions through inpainting and reconstruction tasks~\cite{bar2022visual}. For instance, Painter~\cite{wang2023images} formulates dense prediction tasks as masked image inpainting, demonstrating in-context capabilities across multiple vision tasks. By leveraging pre-trained diffusion models~\cite{rombach2022high,zhao2025diception}, several methods~\cite{wang2023context,geng2024instructdiffusion,qin2024unicontrol,van2024simple} utilize visual or textual instruction to guide generation and enhance adaptability across various tasks. The sequential modeling methods are largely inspired by breakthroughs in large language models and apply the sequence-to-sequence framework to visual data~\cite{touvron2023llama}. For these methods, visual data is typically quantized into sequences of discrete tokens~\cite{van2017neural}. The model is optimized through next-token prediction~\cite{brown2020language}. Recently, \citet{bai2024sequential} introduce a framework that extends this concept to vision without relying on linguistic data, which treats visual data as a ``visual sentence''. By representing images and videos as one-dimensional sequences, this method~\cite{bai2024sequential} enables a unified transformer that can tackle image and video tasks within a single framework, expanding the scope of sequential modeling in computer vision.

In this paper, from the respective of image-resembling generation, we propose a universal diffusion framework with a transformer architecture tailored for visual data, which preserves spatial-temporal structure and minimizes information loss. Trained exclusively on visual data, our flexible framework unifies image and video tasks, advancing toward a generalist model in computer vision.

\myPara{Diffusion transformer.}
By resorting to vision transformer~(ViT)~\cite{alex2021vit,liu2021swin,fan2021multiscale}, recent advancements~\cite{peebles2023scalable,bao2023all,zheng2023fast,gao2023masked,bao2023one,chen2024gentron,crowson2024scalable} in generative modeling achieves significant improvements in scalability and performance for both image~\cite{ramesh2022hierarchical,chen2023pixart,tian2024u,zhuo2024lumina,esser2024scaling} and video generation~\cite{gupta2023photorealistic,polyak2024movie,yang2024cogvideox,ma2024latte}.
Among these advancements, U-ViT~\cite{bao2023all} treats all inputs as tokens by combining transformer blocks with a U-net architecture. DiT~\cite{peebles2023scalable} employs a straightforward and non-hierarchical transformer structure, showcasing the scalability and versatility of diffusion transformers. MDT~\cite{gao2023masked} and MaskDiT~\cite{zheng2023fast} enhance the training efficiency of DiT by using a masking strategy~\cite{he2022masked}. Subsequently, Stable Diffusion~3~\cite{esser2024scaling} introduces a novel transformer-based architecture for text-to-image generation, which enables bidirectional interaction between image and text. Furthermore, diffusion transformers demonstrate robust capabilities for spatial-temporal modeling in video generation~\cite{videoworldsimulators2024}. Previous methods~\cite{ma2024latte,chen2024gentron} utilize separate spatial and temporal attention mechanisms to reduce intensive computational costs. Besides, recent works~\cite{gupta2023photorealistic,polyak2024movie,yang2024cogvideox} have proposed using 3D full attention to capture spatial-temporal information, ensuring consistency for large-moving objects.
While diffusion transformers have shown impressive potential in visual content generation, their capability to serve as a large vision model unifying multiple vision tasks remains underexplored. In this paper, we introduce a new joint diffusion transformer with full-sequence joint attention that effectively integrates diverse vision tasks into a cohesive framework, elevating diffusion transformers to a new level of unified understanding and generation.

\myPara{In-context learning.}
In-context learning is initially conceptualized with GPT-3~\cite{brown2020language}. It has revolutionized the approach to task-specific model training by allowing models to infer and execute tasks based directly on contextual examples provided in prompts~\cite{zhang2023trained}. This paradigm shift enables models to perform complex reasoning and novel pattern recognition without direct training on those specific tasks. Extending beyond text, Flamingo~\cite{alayrac2022flamingo} incorporates visual inputs and broadens in-context learning to multi-modal tasks such as image captioning, visual question answering, and optical character recognition. This demonstrates the model’s ability to integrate and interpret both textual and visual data, enhancing its application across different domains. In the realm of computer vision, the concept of in-context learning is explored through methods such as visual prompting~\cite{bar2022visual}, which infers tasks directly from concatenated image examples and queries. In this paper, we build on this idea. A set of examples are sampled as task definitions and concatenated with the input query for the model, to obtain predictions accordingly.

\begin{figure*}
    \centering
    \includegraphics[width=.94\linewidth]{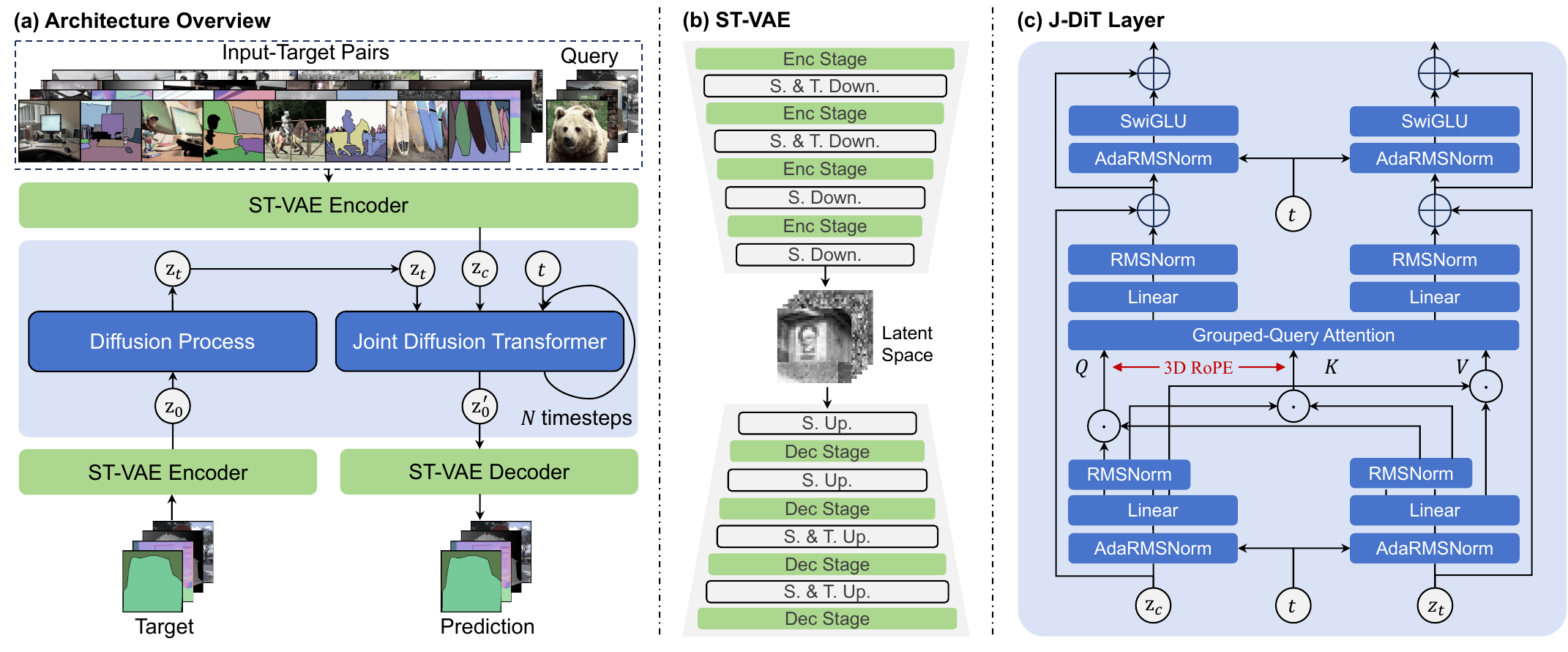}
    \caption{\textbf{Overview of Large Vision Diffusion Model (LaVin-DiT).} 
    As shown in panel (a), the model initially compresses input visual data from the pixel space into a latent space, where multiple input-target pairs serve as the task context. A target is perturbed with Gaussian noise through a diffusion process. Guided by the task context and query, the Joint Diffusion Transformer (J-DiT) iteratively denoises this noisy target over $N$ timesteps to recover a clean latent representation. The prediction is then generated via the ST-VAE decoder.
    Panels (b) and (c) provide architectural details of the ST-VAE and J-DiT, respectively. 
    ``Down.'' and ``Up.'' indicate the downsampling and upsampling, respectively.
    Concatenation is represented by $\odot$.
    }
    \label{fig:overview}
\end{figure*}
\section{Method}
\label{sec:method}

\myPara{Problem setup.} Computer vision includes a series of tasks like object detection~\cite{ren2016faster,lin2017feature,carion2020end} and panoptic segmentation~\cite{kirillov2019panoptic,xiong2019upsnet,cheng2020panoptic}, which are typically handled by specialized models designed for specific input-target mappings~\cite{fu2019geometry}. While effective for single tasks, this specialization restricts model adaptability and scalability across multiple tasks or diverse visual data. To overcome this limitation, we aim to design a \textit{conditional generative framework} that unifies multiple vision tasks within a single cohesive model. Specifically, given a query $\bm{x}$ (\eg, an image or a video), the framework produces the corresponding prediction $\hat{\bm{y}}$ to approximate the target $\bm{y}$ conditioned on a set of input-target pairs $\bm{s}$. These conditioning pairs provide task definitions and guidance, enabling the model to flexibly adapt to different tasks according to the supplied examples. Formally, the objective is to model the conditional distribution $p(\bm{y}|\bm{x},\bm{s})$.

\myPara{Framework overview.} As shown in Figure~\ref{fig:overview}(a), the proposed Large Vision Diffusion Transformer (\textbf{LaVin-DiT}) framework integrates a spatial-temporal variational autoencoder (ST-VAE) with a joint diffusion transformer to unify multiple vision tasks. Given a vision task, \eg, panoptic segmentation, we first sample a set of input-target pairs as the task definition. Afterward, the set and other visual examples are fed into ST-VAE, which are encoded into latent representations. Subsequently, the encoded representations are patchified and unfolded into a sequential format. The set and input visual data form the conditional latent presentation $\bm{z}_c$, while the target is perturbed with random Gaussian noise, yielding a noisy latent representation $\bm{z}_t$. Both $\bm{z}_c$ and $\bm{z}_t$ are then put into the joint diffusion transformer~(J-DiT), which denoises $\bm{z}_t$ to recover a clean latent representation within the shared latent space. Lastly, the recovered latent representation is passed through the ST-VAE decoder to reconstruct the target in raw pixel space. Below we provide a detailed technical exposition of ST-VAE and J-DiT. 

\subsection{LaVin-DiT Modules}

\subsubsection{ST-VAE}
It is computationally demanding to process visual data in raw pixel space~\cite{rombach2022high}. To address this, we propose to use a spatial-temporal variational autoencoder (ST-VAE)~\cite{videoworldsimulators2024,wang2024omnitokenizer,zhao2024cv}. ST-VAE can efficiently compress spatial and temporal information, and encode them from pixel space into compact latent space. As illustrated in Figure~\ref{fig:overview}(b), ST-VAE uses causal 3D convolutions and deconvolutions to compress and reconstruct visual data. It overall includes an encoder, a decoder, and a latent regularization layer. These components are structured into four symmetric stages with alternating 
$2\times$ downsampling and upsampling. The first two stages operate on both spatial and temporal dimensions, while the last stage affects only the spatial dimension, achieving an effective $4 \times 8 \times 8$ compression and reducing computational load. Besides, we apply a Kullback-Leibler (KL) constraint to regularize the Gaussian latent space. 

To prevent future information leakage and its adverse effect on temporal predictions, we pad all locations at the start of the temporal convolution space. Additionally, to support both image and video processing, we treat the first frame of an input video independently, compressing it only spatially to maintain temporal independence. Subsequent frames are compressed along both spatial and temporal dimensions. The encoder of ST-VAE compresses the input to a lower-dimensional latent space, and the reconstruction is achieved through a decoding process. Training the ST-VAE occurs in two stages: we first train on images alone, then jointly on both images and videos. During each stage, we optimize the model using a combination of the mean squared error, perceptual loss~\cite{rombach2022high,zhang2018perceptual}, and adversarial loss~\cite{rombach2022high}.

\subsubsection{J-DiT} 
Diffusion transformers (DiT)~\cite{peebles2023scalable} have emerged as a powerful method for generative modeling. Our joint diffusion transformer (J-DiT) builds upon DiT but introduces modifications to support the task-conditioned generation. A key distinction from the original DiT is our consideration of two conceptually different latent representations. The condition latent representation is clean, while the target latent representation is perturbed by Gaussian noise, resulting in potentially distinct value ranges for the two. To handle the difference and improve alignment between task-specific and visual information, we construct separate patch embeddings for the condition and target latents. Each embedding layer uses a patch size of $2 \times 2$, which allows for tailoring the representations for each latent type. As shown in Figure~\ref{fig:overview}, the sampled timestep $t$, along with the condition and target sequences, is fed into a series of diffusion transformer layers. Building on the MM-DiT \cite{esser2024scaling} architecture, we introduce condition- and target-specific adaptive RMS normalization (AdaRN) to modulate each representation space independently. This is achieved through distinct timestep embeddings for the condition and target within AdaRN layers.

\myPara{Full-sequence joint attention.} Full-sequence joint attention is key in our transformer layers, which processes condition and noisy target sequences together to enhance task-specific alignment. As shown in Figure~\ref{fig:overview}(c), the condition and target sequences are linearly projected, concatenated, and processed by a bidirectional attention module, allowing each to operate in its own space while considering the other. To improve speed and memory efficiency, we replace multi-head attention with grouped-query attention~\cite{ainslie2023gqa}, which groups query heads to share a single set of key-value heads. This approach reduces parameters while retaining expressiveness, closely matching standard multi-head attention performance. Besides, to stabilize training with larger models and longer sequences, we add QK-Norm before query-key dot products to control attention entropy growth.  Following~\cite{team2024gemma}, we also apply sandwich normalization after each attention and FFN layer to maintain activation magnitudes amid residual connections.

\myPara{3D rotary position encoding.} Unlike~\cite{bai2024sequential}, we argue that it is sub-optimal to model visual data as a one-dimensional sequence, because 1D positional embedding is limited in capturing precise spatial-temporal positions. Instead, by treating multiple image-annotation pairs or video clips as a single continuous sequence, we can use 3D Rotary Position Encoding (3D RoPE)~\cite{su2024roformer} to represent spatial-temporal relationships concisely. Then, each location in a video can be expressed by a 3D coordinate. With the introduction of 3D RoPE, we provide a unified and accurate spatial-temporal representation of positional encoding for various vision tasks.

\myPara{Training procedure of J-DiT.} We train J-DiT using flow matching~\cite{lipmanflow} in the latent space. Specifically, given a representation $\bm{z}_0$ and noise $\bm{z}_1\sim\mathcal{N}(0,1)$, flow matching defines a linear interpolation
based forward process: $\bm{z}_t=t\bm{z}_0+(1-t)\bm{z}_1$, where the timestep $t\in[0,1]$. This forward process induces a time-dependent velocity field $v(\bm{z}_t,t)$ that drives the flow along the linear path in the direction of $(\bm{z}_0-\bm{z}_1)$. The velocity field defines an ordinary differential equation (ODE): $\text{d}\bm{z}_t=v(\bm{z}_t,t)\text{d}t$. We employ J-DiT that is parameterized by $\bm{\theta}$, to predict the velocity field that transforms noise into a clean latent representation. The training objective of flow matching is to directly regress the target velocity field, leading to the Conditional Flow Matching (CFM) loss~\cite{lipmanflow}: 
\begin{equation}
    \ell_\text{CFM}=\int_0^1\mathbbm{E}[|v_{\bm{\theta}}(\bm{z}_t,t)-(\bm{z}_0-\bm{z}_1)|_2^2]\text{d}t.
\end{equation}
\myPara{Generation procedure of J-DiT.} Upon completion of J-DiT training, we use it to generate new representations by integrating from the noise distribution toward representation distribution. Specifically, starting from noise $\bm{z}'_1\sim\mathcal{N}(0,1)$ at $t=1$, we integrate the learned J-DiT backward to $t=0$ to obtain a representation $\bm{z}'_0$. For instance, using the Euler method, we discretize the time interval [0,1] to $N$ steps with a negative step size $\Delta t=-1/N$ to indicate backward integration in time. At each step $k=0,1/N,\ldots,(N-1)/N$, we update the time and generated representation as follows:
\begin{align}
t^{(k+1/N)} &= t^{(k)} + \Delta t, \\
\bm{z}^{(k+1/N)} &= \bm{z}^{(k)} + v_{\bm{\theta}}(\bm{z}^{(k)}, t^{(k)}) \Delta t,
\end{align}
where $t^{(0)}=1$, $t^{(1)}=0$, $\bm{z}^{(0)} = \bm{z}'_1$, and $\bm{z}^{(1)} = \bm{z}'_0$.
By iteratively applying these updates, we obtain a new presentation for the following decoding process of ST-VAE.

\subsection{LaVin-DiT Inference}
After completing the training of LaVin-DiT, the model becomes versatile and is ready to be applied across a range of downstream tasks. Specifically, when given a query~(\eg, an image or a video) for any chosen task, we randomly sample a set of input-target pairs that define the task. These pairs, alongside the visual input and a Gaussian noise component, are then fed into the Joint Diffusion Transformer (J-DiT). Within J-DiT, these elements are processed to generate a latent representation. Finally, this latent representation is passed through the ST-VAE decoder, which transforms it into the raw pixel space to produce the desired prediction. To better understand this inference procedure, please refer to Figure~\ref{fig:overview}(a).

\section{Experiments}
\label{sec:exp}
\subsection{Setup}
\myPara{Training data.} To unify multiple computer vision tasks, we construct a large-scale multi-task dataset that encompasses indoor and outdoor environments, spanning real-world and synthetic domains. This dataset comprises approximately 3.2 million unique images \cite{deng2009imagenet,cordts2016cityscapes,zhou2017scene,shao2019objects365,lin2014microsoft} and 0.6 million unique videos \cite{soomro2012ucf101,kay2017kinetics,greff2022kubric}, covering over 20 tasks:
\begin{itemize}
    \item \textit{Image-based tasks}: object detection, instance segmentation, panoptic segmentation, pose estimation, edge extraction, depth estimation, surface normal estimation, inpainting, colorization, image restoration tasks (\eg, de-raining, de-glass blur, and de-motion blur), depth-to-image, and normal-to-image generation.
    \item \textit{Video-based tasks}: frame prediction, video depth estimation, video surface normal estimation, video optical flow estimation, video instance segmentation, depth-to-video, and normal-to-video generation.
\end{itemize}
To overcome the limitations of large-scale annotations for depth and surface normal estimation, we generate pseudo depth and normal maps on ImageNet-1K~\cite{deng2009imagenet} by utilizing Depth-anything V2~\cite{yang2024depth} and Stable-Normal (turbo)~\cite{ye2024stablenormal}, respectively.

\myPara{Implementation details.} We conduct training in two stages, progressively increasing the image resolution. In the first stage, we train at a $256 \times 256$ resolution for 100,000 steps, leveraging DeepSpeed ZeRO-2~\cite{rajbhandari2020zero} optimization and gradient checkpointing to manage memory and computational efficiency. We employ a global batch size of 640 and use an AdamW optimizer~\cite{loshchilov2017decoupled} with a learning rate of 0.0001, betas set to 0.9 and 0.95, and weight decay of 0.01. This setup provides stable training across configurations without the need for a warmup or additional regularization techniques. In the second stage, we upscale the resolution to $512 \times 512$ and continue training for an additional 20,000 steps, while the learning rate is adjusted to 0.00005. Other hyperparameters are retained from the first stage. This two-stage strategy enables efficient scaling, ensuring optimal performance across resolutions. By default, we utilize 20 timesteps ($N=20$) during inference.  All experiments are conducted on 64$\times$NVIDIA A100-80G GPUs.

\begin{table*}[ht]
    \centering
    \caption{
        \textbf{Comparison on foreground segmentation, single object detection, and colorization.} 
        For foreground segmentation and single object detection, we report ``mIoU'' (higher is better). 
        For colorization, we report ``LPIPS'' \cite{zhang2018perceptual} and ``MSE'' (lower is better).
        Note that foreground segmentation and single object detection are \textit{unseen} tasks during our training. 
    }
    \begin{tabular}{l|cccc|cccc|cc}
        \toprule
        \multirow{2}{*}{\textbf{Method}} 
            & \multicolumn{4}{c|}{\textbf{Foreground Segmentation} (mIoU ↑)} 
            & \multicolumn{4}{c|}{\textbf{Single Object Detection} (mIoU ↑)} 
            & \multicolumn{2}{c}{\textbf{Colorization} ↓} \\
        \cmidrule{2-11}
        & Split 1 & Split 2 & Split 3 & Split 4 & Split 1 & Split 2 & Split 3 & Split 4 & MSE & LPIPS \\
        \midrule
        
        MAE~\cite{bar2022visual} & 17.42 & 25.70 & 18.64 & 16.53 & 5.49 & 4.98 & 5.24 & 5.84 & 0.43 & 0.55 \\
        MAE-VQGAN~\cite{bar2022visual} & 27.83 & 30.44 & 26.15 & 24.25 & 24.19 & 25.20 & 25.36 & 25.23 & 0.67 & 0.40 \\
        LVM~\cite{bai2024sequential} & 48.94 & 51.29 & 47.66 & 50.82 & 48.25 & 49.60 & 50.08 & 48.92 & 0.51 & 0.46 \\
        \midrule
        LaVin-DiT & \textbf{67.87} & \textbf{75.80} & \textbf{66.98} & \textbf{66.90} & \textbf{67.85} & \textbf{69.32} & \textbf{68.76} & \textbf{68.88} & \textbf{0.24} & \textbf{0.26} \\
        \bottomrule
    \end{tabular}
    \label{tab:exp1}
\end{table*}

\begin{table}[ht]
    \centering
    \caption{
        \textbf{Comparison on NYU-v2 depth estimation, surface normal estimation and ImageNet inpainting}~\cite{silberman2012indoor,deng2009imagenet}. 
        For depth estimation, we report absolute relative difference (AbsRel) and threshold accuracy ($\delta_1$). 
        For surface normal estimation, we report mean angular error (MAE) and angle accuracy within a threshold ($<11.25^\circ$). 
        We report FID for inpainting.
        $\dagger$ denotes evaluations on the official 7B model released by~\cite{bai2024sequential}.}
    \vspace{-0.1cm}
    \resizebox{\columnwidth}{!}{
    \begin{tabular}{l|cc|cc|c}
        \toprule
        \multirow{2}{*}{\textbf{Method}} 
            & \multicolumn{2}{c|}{\textbf{Depth Estimation}} 
            & \multicolumn{2}{c|}{\textbf{Normal Estimation}}
            & \textbf{Inpainting} \\
        \cmidrule{2-6}
        & AbsRel (↓) & $\delta_1$ (↑) & MAE (↓) & $<11.25^\circ$ (↑) & FID (↓) \\
        \midrule
        DPT~\cite{ranftl2021vision} & 9.8 & 90.3 & - & - & - \\
        StableNormal~\cite{ye2024stablenormal} & - & - & 19.707 & 53.042 & - \\
        Marigold~\cite{ke2024repurposing} & 6.0 & 95.9 & 20.864 & 50.457 & - \\
        \midrule
        LVM$\dagger$~\cite{bai2024sequential} & 30.2 & 52.3 & 23.433 & 44.836 & 4.05 \\
        LaVin-DiT & \textbf{6.2} & \textbf{96.1} & \textbf{15.901} & \textbf{58.382} & \textbf{1.65} \\
        \bottomrule
    \end{tabular}}
    \label{tab:exp2}
\end{table}

\myPara{Evaluation protocols.}
We assess our model on a comprehensive range of computer vision tasks spanning both image and video domains. Following established protocols, we report standard metrics for each task.

\subsection{Main Results}
\myPara{Quantitative analysis.}
To assess the effectiveness of our proposed method, we conduct extensive experiments across a broad range of computer vision tasks and report results of the 3.4B model by default, as summarized in Tables \ref{tab:exp1} and \ref{tab:exp2}. Our method consistently outperforms existing baselines across multiple tasks, including challenging cases such as unseen foreground segmentation and single-object detection, demonstrating superior generalization and adaptability across diverse scenarios. Note that unless otherwise specified, we report LaVin-Dit~(3.4B) performance.

As shown in Table \ref{tab:exp1}, we report the performance on foreground segmentation and single object detection across different splits. Our LaVin-DiT achieves significant improvements over baseline methods in all splits. Specifically, in the foreground segmentation task, we attain mIoUs of 67.87\%, 75.80\%, 66.98\%, and 66.90\% across four splits, consistently outperforming previous methods such as LVM~\cite{bai2024sequential} and MAE-VQGAN~\cite{bar2022visual} by a substantial margin. Additionally, for single object detection, our model demonstrates strong performance, achieving top results in all splits. Notably, we achieve a mIoU of 68.88\% in Split 4, which is a considerable margin of 19.96\% over the best-performing baseline LVM. These significant gains highlight our model’s ability to effectively segment and detect objects across a range of scenarios, even when faced with tasks unseen during training. Following prior work~\cite{bar2022visual,bai2024sequential}, we further evaluate our model in the colorization task, where lower LPIPS and MSE values indicate superior performance. As shown in Table~\ref{tab:exp1}, our method achieves an LPIPS of 0.26 and an MSE of 0.24, significantly outperforming all baselines. These results underscore our model’s capability to generate realistic and natural colors from grayscale images, which is essential in restoration and artistic fields. 

To validate the ability of our model to understand the geometric structure of 3D scenes, we evaluate it on NYU-v2 depth estimation and surface normal estimation tasks~\cite{silberman2012indoor}, as shown in Table~\ref{tab:exp2}. As \citet{bai2024sequential} do not report related results in their paper, we conduct evaluations using their official 7B model\footnote{\url{https://huggingface.co/Emma02/LVM_ckpts}}. For depth estimation, our model achieves an AbsRel of 6.2 and a threshold accuracy $\delta_1$ of 96.1\%, demonstrating competitive performance compared to expert models such as Marigold~\cite{ke2024repurposing} and DPT~\cite{ranftl2021vision}. In the surface normal estimation task, our method achieves an MAE of 15.901 and accuracy within a $<11.25^\circ$ threshold of 58.382, surpassing the powerful expert model StableNormal~\cite{ye2024stablenormal}. This performance underscores our model’s proficiency in estimating surface orientations accurately, enhancing its applicability in tasks requiring precise geometrical understanding, such as augmented reality and 3D reconstruction. These results reflect our model’s capability to comprehend the geometric structure of 3D scenes with precision, even in complex environments, which is crucial for real-world applications like 3D scene reconstruction and spatial perception. Furthermore, we compare our LaVin-DiT to LVM on the inpainting task. Using 2,500 randomly selected images from the ImageNet-1K validation set, our model achieves an FID of 1.65, which greatly improves over the FID of 4.05 obtained by LVM.

\begin{figure*}[t]
    \centering
    \includegraphics[width=.935\linewidth]{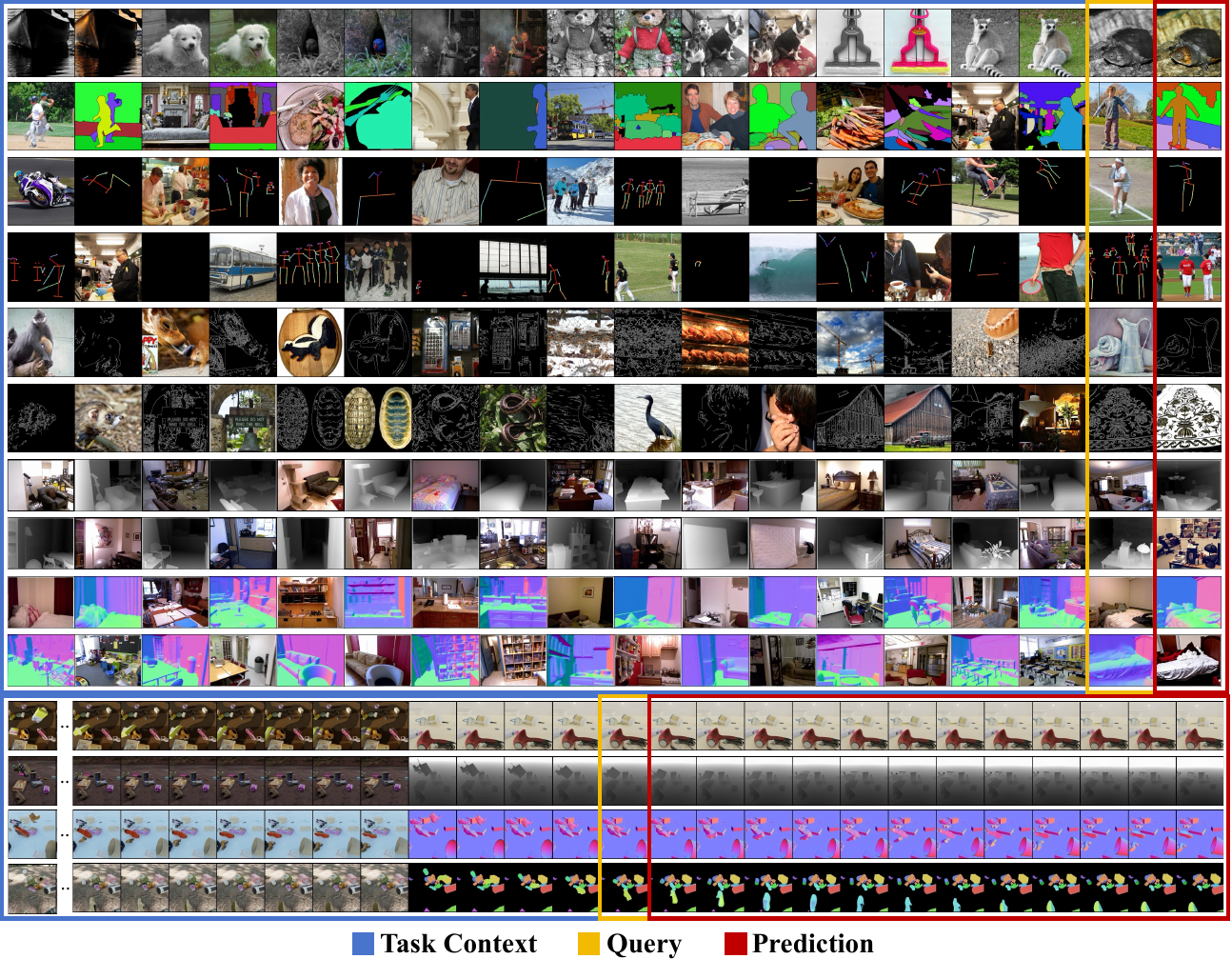}
    \caption{\textbf{Qualitative results on diverse image and video-based tasks.} The first ten rows show image-based tasks, where each row contains a sequence of images interleaved with annotations, followed by a query. The last image is predicted by the model (marked in red). The last four rows show video-based tasks, where each row includes a video sequence with a series of target frames as task context, followed by a query frame. A set of frames in the red box indicates the model’s predictions. \textit{Best viewed in color.}}
    \label{fig:vis}
\end{figure*}
\myPara{Qualitative analysis.}
As shown in Figures \ref{fig:vis}, we present qualitative results in a wide variety of image-based and video-based tasks. Our model consistently follows task contexts and precisely generates the corresponding predictions. Furthermore, given sequential frames with task contexts, our model generates predictions for the subsequent 12 frames, which exhibits its ability to handle temporal consistency and scene dynamics effectively.

\begin{figure*}[ht]
    \centering
    \begin{minipage}[!t]{0.32\linewidth}
        \centering
        \includegraphics[width=.95\linewidth]{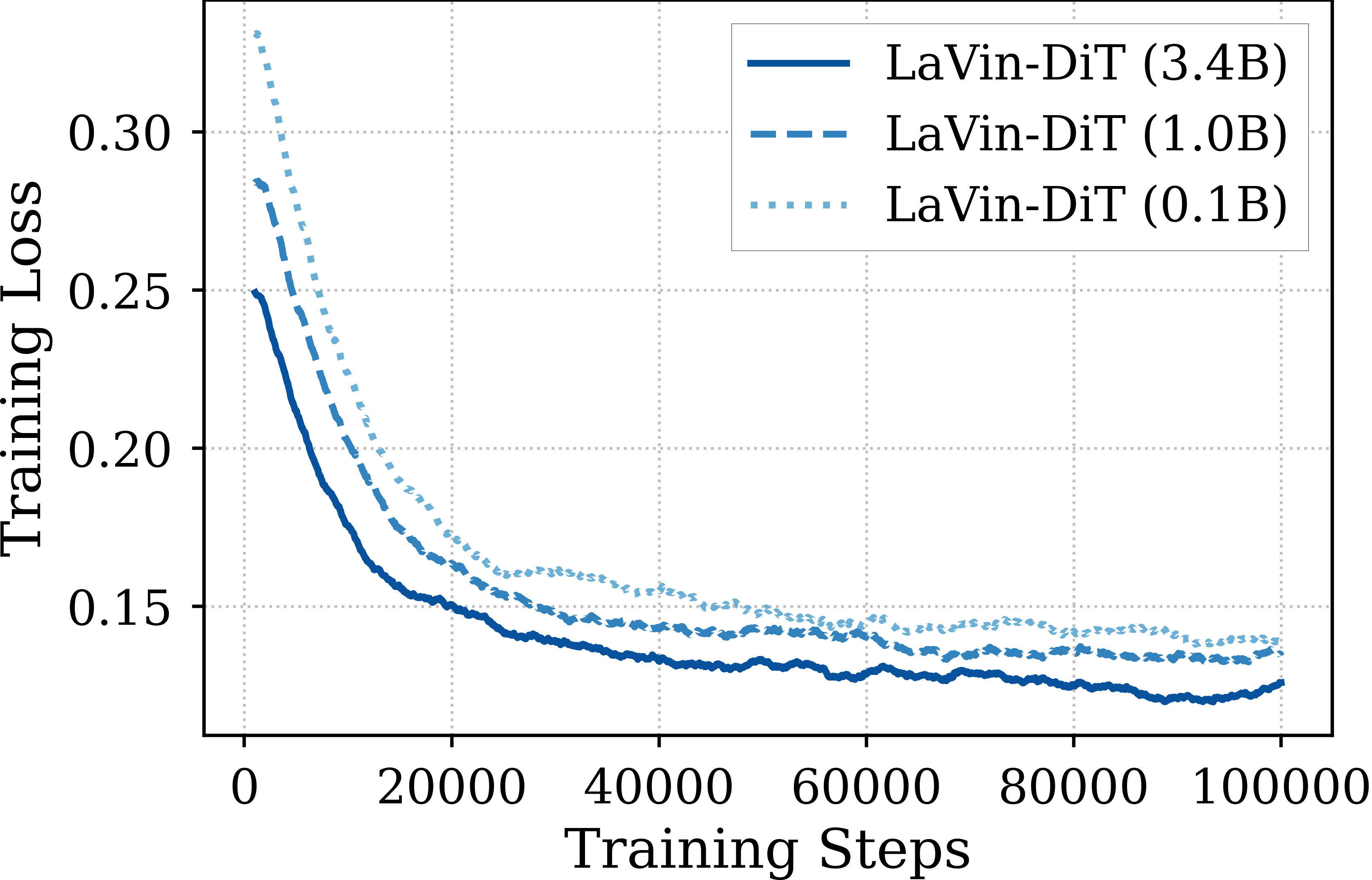}
        \caption{\textbf{Training loss curves for LaVin-DiT of varying model sizes.} 
        The 3.4B model demonstrates faster convergence, achieving lower training losses than smaller models as training progresses.}
        \label{fig:loss_curve}
    \end{minipage}
    \hfill
    \centering
    \begin{minipage}[!t]{0.32\linewidth}
        \centering
        \includegraphics[width=.95\linewidth]{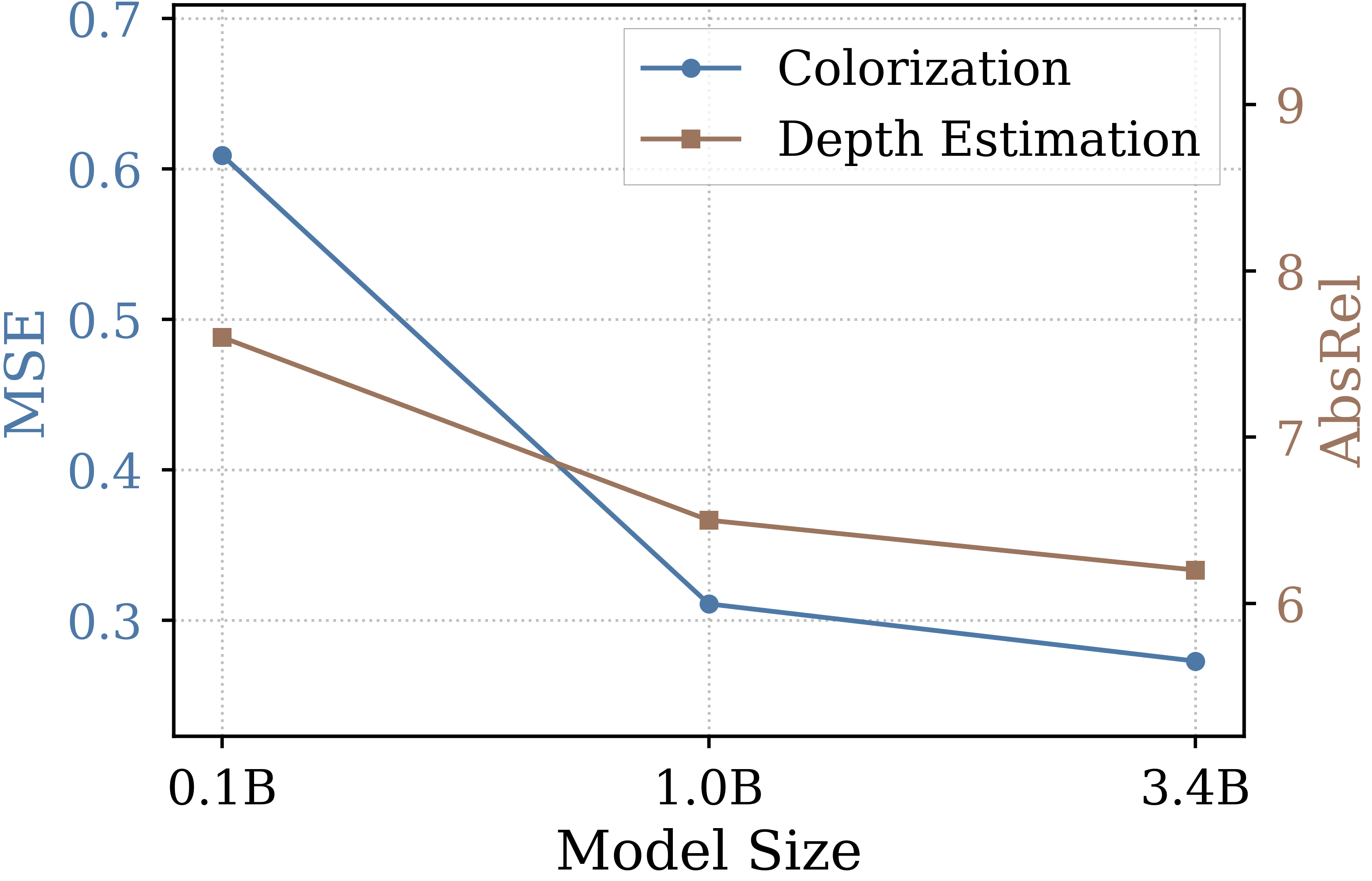}
        \caption{\textbf{Performance for LaVin-DiT of varying sizes.} Comparison of LaVin-DiT with different parameters on colorization (MSE) and depth estimation (AbsRel). Lower values indicate better performance.}
        \label{fig:metric_curve}
    \end{minipage}
    \hfill
    \begin{minipage}[!t]{0.32\linewidth}
        \centering
        \includegraphics[width=.95\linewidth]{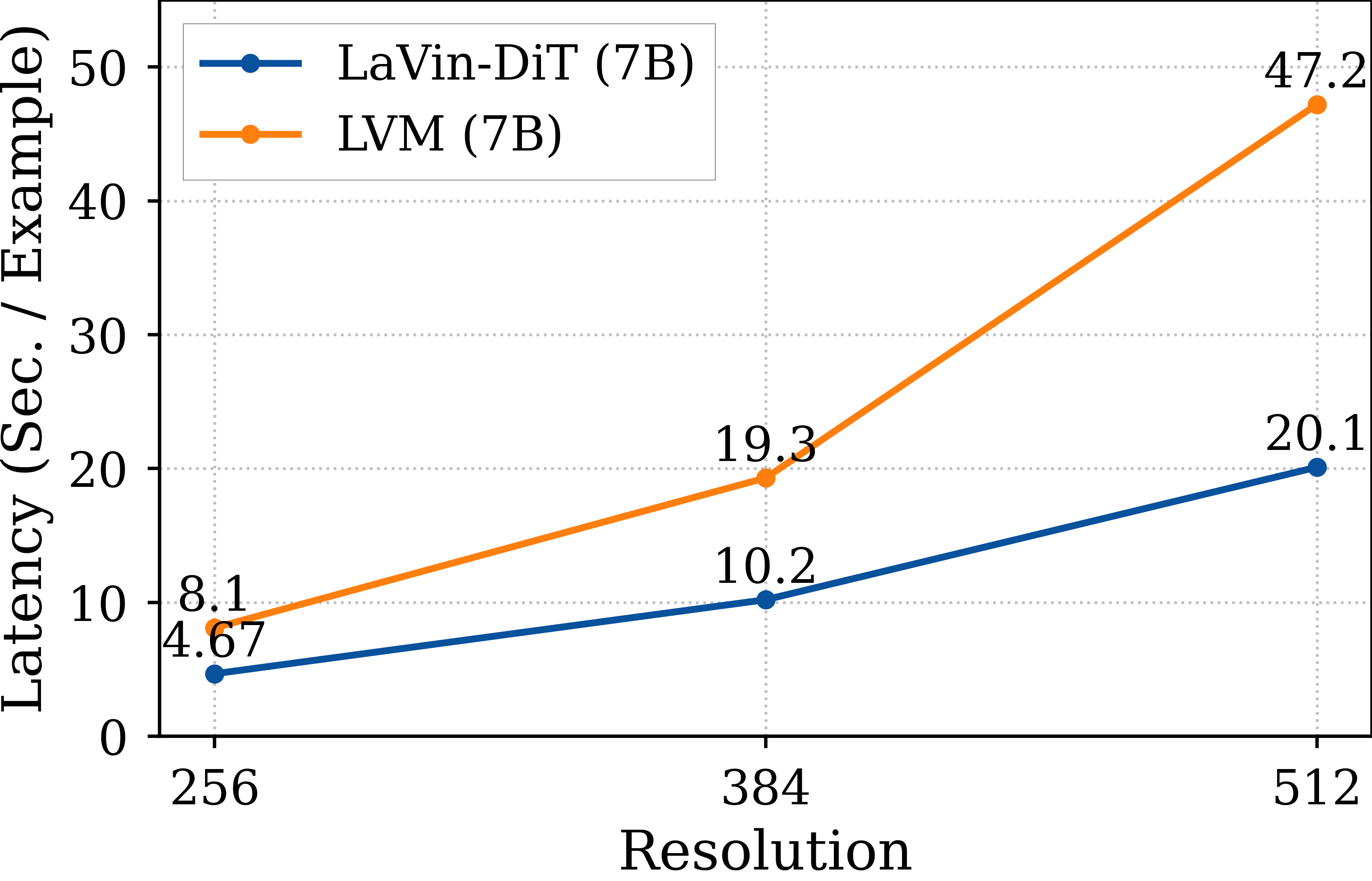}
        \caption{\textbf{Inference latency comparison.} LaVin-DiT consistently achieves lower latency than LVM \cite{bai2024sequential} across different resolutions, as tested on an A100-80G GPU with 8 input-target pairs.}
        \label{fig:latency_comparison}
    \end{minipage}
    \hfill
\end{figure*}

\begin{figure*}[ht]
    \centering
    \includegraphics[width=.951\linewidth]{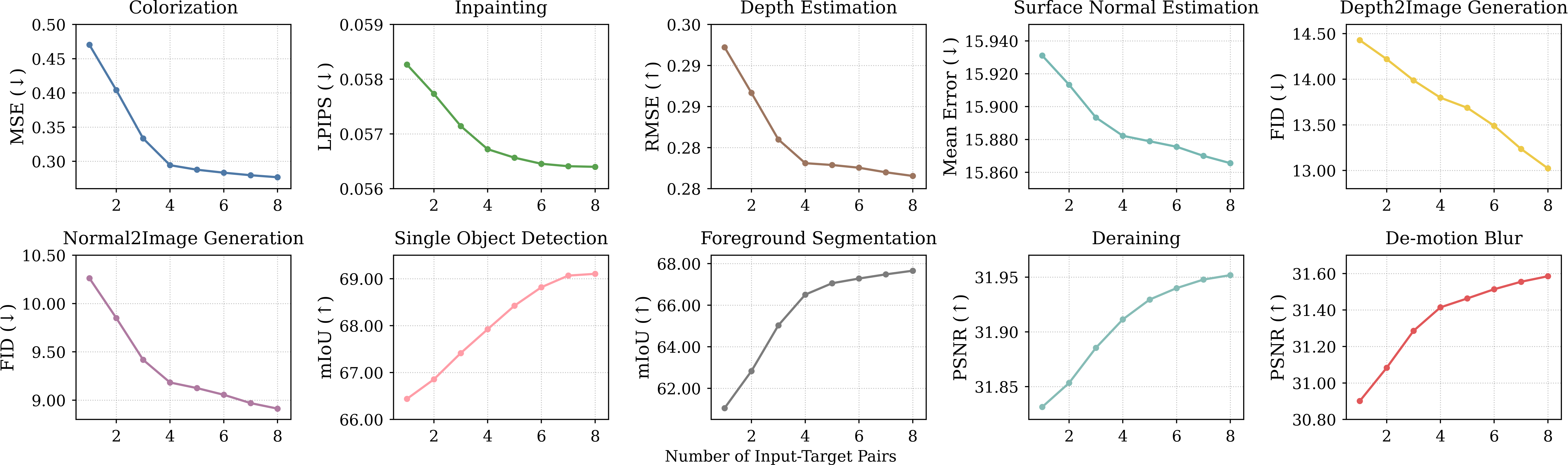}
    \caption{\textbf{Effect of task context length.} Longer task context can consistently improve the performance of downstream tasks.}
    \label{fig:context_length}
\end{figure*}

\subsection{Scalability}
To investigate the scalability of the proposed LaVin-DiT, we conduct experiments with three model sizes, \ie, 0.1B, 1.0B, and 3.4B parameters. We train the three models for 100,000 steps. Figure~\ref{fig:loss_curve} illustrates the training loss curves, which shows that larger models consistently achieve lower loss values. Additionally, the 3.4B model converges more rapidly, reaching smaller loss values in fewer training steps. This accelerated convergence suggests that larger models are better equipped to capture complex data patterns, leading to improved learning efficiency. The observed training dynamics underscore the advantages of scaling up model capacity for complex vision tasks, where larger models can more effectively capture diverse data characteristics.

Beyond training dynamics, the model size also has a substantial impact on downstream task performance. This is evident in colorization and depth estimation tasks, which were selected for their distinct requirements in capturing color fidelity and spatial structure. As seen in Figure~\ref{fig:metric_curve}, model performance improves consistently as its scale increases. Specifically, for colorization, the 3.4B model achieves an MSE of 0.273, significantly outperforming the 1.0B and 0.1B models that achieve MSEs of 0.311 and 0.609, respectively. Similarly, in depth estimation, the 3.4B model attains an AbsRel of 6.2, compared to 6.5 and 7.6 for the 1.0B and 0.1B models. These results demonstrate that larger models indeed deliver enhanced performance across multiple tasks, affirming LaVin-DiT as a scalable and adaptable framework for high-performance vision applications.

\subsection{Inference Latency Analysis}
As demonstrated in Figure~\ref{fig:latency_comparison}, we compare the inference latency of LaVin-DiT and LVM (both 7B models) across increasing resolutions, demonstrating that our method is consistently more efficient. At a resolution of 256, LaVin-DiT requires only 4.67 seconds per example, while LVM takes 8.1 seconds, with this efficiency gap widening at higher resolutions (\eg, 20.1 seconds \textit{v.s.} 47.2 seconds at 512). This difference underscores a key advantage of diffusion models for vision tasks: unlike autoregressive models that process tokens sequentially and become increasingly time-intensive with larger inputs, diffusion models process tokens in parallel, allowing them to scale more effectively. This parallelism makes our LaVin-DiT a more suitable choice for large-scale vision applications.

\subsection{Effect of Task Context Length}
In-context learning enables the model to adapt to new tasks using a few examples, with performance generally improving as more examples are provided. We investigate this by assessing the effect of task context length across ten downstream tasks. As shown in Figure~\ref{fig:context_length}, the model consistently benefits from longer task contexts, achieving notable performance gains. For instance, with more input-target pairs, LaVin-DiT achieves lower FID in depth-to-image generation and higher PSNR in de-motion blur tasks. These results demonstrate that LaVin-DiT effectively leverages extended task context, highlighting its capacity to utilize additional information for enhanced task adaptation and accuracy.

\vspace{-0.05cm}
\section{Conclusion}
\label{sec:conclusion}
We present LaVin-DiT, a scalable and unified foundation model for computer vision that integrates a spatial-temporal variational autoencoder and a diffusion transformer to efficiently process high-dimensional vision data while preserving spatial and visual coherence. Through in-context learning, LaVin-DiT adapts effectively to a wide range of tasks without fine-tuning, which shows remarkable versatility and adaptability. Extensive experiments validate LaVin-DiT’s scalability and performance, positioning it as a promising framework for developing generalist vision models.

\myPara{Limitations.}
Despite its advantages, LaVin-DiT is limited by current constraints in large-scale training data, diverse task annotations, and computational resources, especially in comparison to large language models. While our model achieves strong results on seen tasks and related unseen tasks, it struggles with generalization when task definitions deviate significantly from the training distribution. This limitation highlights a key challenge in developing vision models that can generalize effectively to entirely new tasks defined solely by task context.

\myPara{Future work.}
Future research should explore scaling LaVin-DiT further in terms of model capacity, dataset diversity, and task complexity to push the boundaries of vision generalization. We anticipate that as these elements expand, LaVin-DiT and similar models may gain the ability to handle arbitrary (out-of-training) vision tasks, guided only by a few input-target pairs. Additionally, investigating methods to select optimal task context automatically could provide a rapid and effective pathway to enhance model performance, ensuring that it leverages the most relevant examples for each task. These directions will drive further advances in developing robust, adaptable, and highly generalized foundation models for computer vision.


{
    \small
    \bibliographystyle{ieeenat_fullname}
    \bibliography{main}
}
\clearpage
\maketitlesupplementary
\appendix


\section{More Technical Details of LaVin-DiT}

\subsection{Details of 3D RoPE}\label{appendix:3d_rope_details}
Recall that we represent task context and query as a unified sequence of frames, which is a 3D representation. Afterward, we extend RoPE from 1D to 3D format to capture the essential structure of visual data.
Specifically, each token in an input sequence is associated with a 3D coordinate $(t, x, y)$, representing its position in temporal and spatial dimensions. The 3D RoPE encodes positional information by decomposing it into three separate 1D RoPEs along the temporal and spatial axes, allowing the model to capture relative positional relationships across all dimensions inherently.

Technically, for each axis $a \in \{ t, x, y \}$, we define a rotation matrix $R_p^{(a)}$ that operates on a dedicated subspace of an embedding vector $z$. The embedding vector is partitioned accordingly: $z = [z^{(t)}, \; z^{(x)}, \; z^{(y)}]$, where each subvector $z^{(a)} \in \mathbb{R}^{d_a}$ corresponds to axis $a$ and $d = d_t + d_x + d_y$. The rotation matrix $R_p^{(a)}$ is constructed in a block-wise manner, rotating each pair of dimensions $(2i, 2i+1)$ by an angle $\theta_{p,i}^{(a)} = p^{(a)} \cdot \omega_i^{(a)}$, where $\omega_i^{(a)} = \omega_{\text{base}}^{-2i/d_a}$ and $\omega_{\text{base}}$ is a predefined constant:
\begin{align}
R_p^{(a)} &= 
\begin{bmatrix}
R_p^{(a,0)} & & \\
& \ddots & \\
& & R_p^{(a,d_a/2 - 1)}
\end{bmatrix}, \quad \text{where} \\
R_p^{(a,i)} &= 
\begin{bmatrix}
\cos\left(\theta_{p,i}^{(a)}\right) & -\sin\left(\theta_{p,i}^{(a)}\right) \\
\sin\left(\theta_{p,i}^{(a)}\right) & \cos\left(\theta_{p,i}^{(a)}\right)
\end{bmatrix}.
\end{align}
When computing self-attention, the rotated query $q$ and key $k$ are obtained by applying the rotation matrices: $q^{\prime(a)} = R_p^{(a)} q^{(a)}$ and $k^{\prime(a)} = R_p^{(a)} k^{(a)}$. The full rotated query and key are then $q^{\prime} = [q^{\prime(t)}, \; q^{\prime(x)}, \; q^{\prime(y)}]$ and $k^{\prime} = [k^{\prime(t)}, \; k^{\prime(x)}, \; k^{\prime(y)}]$. When computing the attention between tokens at positions $j$ and $k$, the dot product incorporates the rotations from all axes:
\begin{equation}
(q_j^{\prime\top}) k^{\prime}_k = \sum_{a \in \{ t, x, y \}} \left( q^{(a)} \right)^\top R_j^{(a)\top} R_k^{(a)} k^{(a)}.
\end{equation}
The key property of rotation matrices is that the product of two rotation matrices corresponds to a rotation by the difference of their angles:
\begin{equation}
R_j^{(a)\top} R_k^{(a)} = R_{j - k}^{(a)},
\end{equation}
where $R_{p - q}^{(a)}$ is the rotation matrix for the relative position $j^{(a)} - k^{(a)}$, constructed as:
\begin{align}
R_{j - k}^{(a)} &= 
\begin{bmatrix}
R_{j - k}^{(a,0)} & & \\
& \ddots & \\
& & R_{j - k}^{(a, N_a - 1)}
\end{bmatrix}, \quad \text{where} \\
R_{j - k}^{(a,i)} &= 
\begin{bmatrix}
\cos\left(\Delta_{jk}^{(a)} \omega_i^{(a)}\right) & -\sin\left(\Delta_{jk}^{(a)} \omega_i^{(a)}\right) \\
\sin\left(\Delta_{jk}^{(a)} \omega_i^{(a)}\right) & \cos\left(\Delta_{jk}^{(a)} \omega_i^{(a)}\right)
\end{bmatrix}, \\ \Delta_{jk}^{(a)} &= j^{(a)} - k^{(a)}.
\end{align}
This block-wise matrix format explicitly shows that the attention score depends on the relative positions $j^{(a)} - k^{(a)}$ along each axis $a$.

\subsection{Algorithm Flows of LaVin-DiT}\label{appendix:alg_flows}
In this section, we present algorithm flows of the proposed LaVin-DiT. It is built upon the flow matching framework~\cite{lipmanflow}. The training and inference procedures are provided in Algorithm~\ref{algo:train} and Algorithm~\ref{algo:sample}, respectively.

\begin{algorithm}[H]
\caption{LaVin-DiT Training Procedure}
\begin{algorithmic}[1]
\Require ST-VAE encoder $\operatorname{Enc}(\cdot)$, dataset $\mathcal{D} = \{ \bm{x}_i \}_{i=1}^K$, initialized parameters $\bm{\theta}$ of vector field $v_{\bm{\theta}}(\bm{z}, t)$, total iterations $T$, learning rate $\eta$.
\For{$n = 1$ to $T$}
\State Sample $\bm{x} \sim \mathcal{D},\ \bm{c} \sim \mathcal{D}$
\State Compute latents: $\bm{z}_0 \leftarrow \operatorname{Enc}(\bm{x}),\ \bm{z}_c \leftarrow \operatorname{Enc}(\bm{c})$
\State Initialize random latent: $\bm{z}_1 \sim \mathcal{N}(0, 1)$
\State Sample time step: $t \sim \operatorname{LogitNormal}(0, 1)$
\State Interpolate: $\bm{z}_t \leftarrow (1 - t) \bm{z}_1 + t \bm{z}_0$
\State Target vector: $\bm{u} \leftarrow \bm{z}_0 - \bm{z}_1$
\State Predicted vector: $\bm{v} \leftarrow v_{\bm{\theta}}(\bm{z}_t,\ \bm{z}_c,\ t)$
\State Compute loss: $\mathcal{L} \leftarrow \mathbb{E}[|\bm{v} - \bm{u}|_2^2]$
\State Update parameters: $\bm{\theta} \leftarrow \bm{\theta} - \eta \nabla_{\bm{\theta}} \mathcal{L}$
\EndFor
\end{algorithmic}
\label{algo:train}
\end{algorithm}

\myPara{Training procedure.}
As illustrated in Algorithm~\ref{algo:train}, the primary goal is to learn a vector field $v_{\bm{\theta}}(\bm{z}, t)$ that maps the latent space dynamics conditioned on the target latent $\bm{z}_0$, the task context latent $\bm{z}_c$, and a time step $t$. The training process iteratively refines the parameters $\bm{\theta}$ to minimize the discrepancy between the predicted and ground-truth latent trajectories.

\myPara{Inference procedure.}
This process, described in Algorithm~\ref{algo:sample}, employs the learned vector field $v_{\bm{\theta}}$ to sample in the latent space. Starting with an initial latent $\bm{z}_1 \sim \mathcal{N}(0, 1)$, the method denoises iteratively using the Euler method.

\begin{algorithm}[th]
\caption{LaVin-DiT Inference Procedure}
\begin{algorithmic}[1]
\Require Trained vector field $v_{\bm{\theta}}(z, t)$, ST-VAE encoder $\operatorname{Enc}(\cdot)$, ST-VAE decoder $\operatorname{Dec}(\cdot)$, timesteps $N$, dataset $\mathcal{D} = \{ \bm{x}_i \}_{i=1}^K$.
\State Set step size $\Delta t \leftarrow \frac{1}{N}$, initialize $t^{(N)} \leftarrow 1$
\State Sample initial latent: $\bm{z}_1 \sim \mathcal{N}(0, 1)$
\State Encode condition: $\bm{z}_c \leftarrow \operatorname{Enc}(\bm{c}),\ \bm{c} \sim \mathcal{D}$
\For{$k = N$ down to $1$}
\State Update time: $t^{(k-1)} \leftarrow t^{(k)} - \Delta t$
\State Compute vector field: $\bm{v}^{(k)} \leftarrow v_{\bm{\theta}}(\bm{z}^{(k)},\ \bm{z}_c,\ t^{(k)})$
\State Update latent: $\bm{z}^{(k-1)} \leftarrow \bm{z}^{(k)} - \Delta t \cdot \bm{v}^{(k)}$
\EndFor
\State Decode sample: $\hat{\bm{y}} \leftarrow \operatorname{Dec}(\bm{z}_0)$
\end{algorithmic}
\label{algo:sample}
\end{algorithm}

\begin{table}[!ht]
\centering
\caption{Configurations of LaVin-DiT with different numbers of parameters.}
\begin{tabular}{lccc}
\toprule
 & \multicolumn{3}{c}{\textbf{LaVin-DiT}}  \\
 & \textbf{0.1B} & \textbf{1.0B} & \textbf{3.4B} \\
\hline
\textbf{Latent channels} & 16 & 16 & 16 \\
\textbf{Patch size} & $2 \times 2$ & $2 \times 2$ & $2 \times 2$ \\
\textbf{Hidden channels} & 512 & 1024 & 2304 \\
\textbf{Num. layers} & 12 & 28 & 22 \\
\textbf{Num. heads} & 8 & 16 & 32 \\
\textbf{K.V. groups} & - & - & 4 \\
\textbf{Drop path} & 0.0 & 0.1 & 0.1 \\
\textbf{Uncond. ratio} & 0.1 & 0.1 & 0.1 \\
\textbf{Grad. clip} & 1.0 & 1.0 & 1.0 \\
\textbf{EMA moment.} & 0.9999 & 0.9999 & 0.9999 \\
\textbf{Extra norm.} & - & S-Norm. & S-Norm. \\
\textbf{Position embed.} & 3D-RoPE & 3D-RoPE & 3D-RoPE \\
\bottomrule
\end{tabular}%
\label{tab:model_detail}
\end{table}

\section{Supplementary Experimental Settings}

\subsection{Large-Scale Multi-Task Dataset Composition}\label{appendix:dataset_details}
Recall that we build a large-scale multi-task dataset to unify diverse computer vision tasks. We integrate multiple public image-level and video-level task benchmarks into a large-scale dataset for training. Details are listed in Table~\ref{tab:dataset_card}.

\begin{table*}[th]
    \centering
    \caption{Summary of the large-scale multi-task dataset used in LaVin-DiT, including the number of examples and annotation types for each component dataset. Tasks range from visual understanding and generation.}
    \begin{tabular}{p{5cm}|p{4cm}|p{3cm}|p{3.5cm}}
    \toprule
    \textbf{Task} & \textbf{Dataset} & \textbf{Number of Samples} & \textbf{Annotation Type} \\
    \hline
    \multirow{2}{=}{Single Object Detection} 
        & COCO 2017 train~\cite{lin2014microsoft} & 117,266 & Ground Truth \\
        & Object365 train~\cite{shao2019objects365} & 1,728,778 & Ground Truth \\
    \hline
    \multirow{3}{=}{Instance Segmentation} 
        & COCO 2017 train~\cite{lin2014microsoft} & 117,266 & Ground Truth \\
        & ADE20K train+val~\cite{zhou2017scene} & 19,020 & Ground Truth \\
        & Cityscapes train+val~\cite{cordts2016cityscapes} & 3,457 & Ground Truth \\
    \hline
    \multirow{3}{=}{Panoptic Segmentation} 
        & COCO 2017 train~\cite{lin2014microsoft} & 117,266 & Ground Truth \\
        & ADE20K train+val~\cite{zhou2017scene} & 19,020 & Ground Truth \\
        & Cityscapes train+val~\cite{cordts2016cityscapes} & 3,457 & Ground Truth \\
    \hline
    Pose Estimation & COCO 2017 train~\cite{lin2014microsoft} & 64,115 & Ground Truth \\\hline
    Pose-to-Image Generation & COCO 2017 train~\cite{lin2014microsoft}  & 64,115  & Ground Truth  \\
    \hline
    Depth Estimation & ImageNet1K train~\cite{deng2009imagenet} & 1,281,167 & Depth-anything V2 \\\hline
    Depth-to-Image Generation & ImageNet1K train~\cite{deng2009imagenet} & 1,281,167 & Depth-anything V2\\
    \hline
    \multirow{3}{=}{Surface Normal Estimation} & COCO 2017 train~\cite{lin2014microsoft} & 117,266 & Stable-Normal (turbo) \\
    & ADE20K train+val~\cite{zhou2017scene} & 19,020 & Stable-Normal (turbo) \\
        & Cityscapes train+val~\cite{cordts2016cityscapes} & 3,457 & Stable-Normal (turbo) \\
    \hline
     \multirow{3}{=}{Normal-to-Image Generation} & COCO 2017 train~\cite{lin2014microsoft} & 117,266 & Stable-Normal (turbo) \\
     & ADE20K train+val~\cite{zhou2017scene} & 19,020 & Stable-Normal (turbo) \\
        & Cityscapes train+val~\cite{cordts2016cityscapes} & 3,457 & Stable-Normal (turbo) \\
    \hline
    \multirow{2}{=}{Edge Detection}
        & ImageNet1K~\cite{deng2009imagenet} train & 1,281,167 & Canny (OpenCV) \\
        & COCO 2017 train~\cite{lin2014microsoft} & 117,266 & Canny (OpenCV) \\
    \hline
    \multirow{2}{=}{Inpainting} 
        & ImageNet1K train~\cite{deng2009imagenet} & 1,281,167 & Crop (OpenCV) \\
        & COCO 2017 train~\cite{lin2014microsoft} & 117,266 & Crop (OpenCV) \\
    \hline
    \multirow{2}{=}{Colorization} 
        & ImageNet1K train~\cite{deng2009imagenet} & 1,281,167 & Grayscale (OpenCV) \\
        & COCO 2017 train~\cite{lin2014microsoft} & 117,266 & Grayscale (OpenCV) \\
    \hline
    \multirow{2}{=}{De-glass Blur} 
        & ImageNet1K train~\cite{deng2009imagenet} & 1,281,167 & Albumentations \\
        & COCO 2017 train~\cite{lin2014microsoft} & 117,266 & Albumentations \\
    \hline
    \multirow{2}{=}{De-motion Blur} 
        & ImageNet1K train~\cite{deng2009imagenet} & 1,281,167 & Albumentations \\
        & COCO 2017 train~\cite{lin2014microsoft} & 117,266 & Albumentations \\
    \hline
    \multirow{2}{=}{De-raining} 
        & ImageNet1K train~\cite{deng2009imagenet} & 1,281,167 & Albumentations \\
        & COCO 2017 train~\cite{lin2014microsoft} & 117,266 & Albumentations \\
    \hline
    \multirow{3}{=}{Frame Prediction} 
        & UCF101 train~\cite{soomro2012ucf101} & 7,629 & N/A \\
        & Kinetic 700 train+val~\cite{kay2017kinetics} & 570,465 & N/A \\
        & Kubric train~\cite{greff2022kubric} & 48,689 & N/A \\
    \hline
    Video Depth Estimation & Kubric train~\cite{greff2022kubric} & 48,689 & Ground Truth \\
    \hline
    Depth-to-Video Generation & Kubric train~\cite{greff2022kubric} & 48,689 & Ground Truth \\
    \hline
    Video Surface Normal Estimation & Kubric train~\cite{greff2022kubric} & 48,689 & Ground Truth \\
    \hline
    Normal-to-Video Generation & Kubric train~\cite{greff2022kubric} & 48,689 & Ground Truth \\
    \hline
    Video Optical Flow Estimation 
        & Kubric train~\cite{greff2022kubric} & 48,689 & Ground Truth \\
    \hline
    Video Instance Segmentation 
        & Kubric train~\cite{greff2022kubric} & 48,689 & Ground Truth \\
    \bottomrule
    \end{tabular}
    \label{tab:dataset_card}
\end{table*}

\subsection{Evaluation Metrics}\label{appendix:metrics}
In this work, we provide quantitative results for 10 tasks (The others are presented with visualization results).  Here we introduce the evaluation metrics for these 10 tasks. 

\myPara{Colorization.} We randomly sample 1,000 images from ImageNet-1K validation set~\cite{deng2009imagenet} and convert them into grayscale. We adopt LPIPS~\cite{zhang2018perceptual} and mean squared error (MSE) as metrics.

\myPara{Inpainting.} We randomly sample 1,000 images from ImageNet-1K validation set~\cite{deng2009imagenet} and mask out a $128 \times 128$ region for each image. We adopt the LPIPS~\cite{zhang2018perceptual} and Frechet Inception Distance (FID) as metrics.

\myPara{Depth Estimation.} We evaluate our model on NYUv2 test set~\cite{silberman2012indoor}, including 654 images. Following the protocol of affine-invariant depth evaluation~\cite{ranftl2020towards}, we first align the prediction to the ground truth with the least squares fitting. Afterwards, we adopt Absolute Mean Relative Error~(AbsRel) and Mean Squared Error~(MSE) as metrics.

\myPara{Surface Normal Estimation.} We evaluate our model on NYUv2 test set~\cite{silberman2012indoor}. Following the protocol used in~\cite{bae2024rethinking}, we calculate the angular error between the prediction and the ground-truth normal maps and use the mean angular error as the metric.

\myPara{Depth-to-Image Generation.} We adopt all samples in the NYUv2 dataset~\cite{silberman2012indoor}, including 1,449 images. Given the pseudo label generated via Depth-anything V2 or Stable-Normal (turbo), we generate the corresponding RGB image and use the LPIPS~\cite{zhang2018perceptual} and Frechet Inception Distance (FID) as metrics.

\myPara{Normal-to-Image Generation.} The metrics are the same those in Depth-to-Image Generation. 

\myPara{Single Object Detection.}  We evaluate the model on the Pascal-5i dataset~\cite{shaban2017one} and adopt the mean intersection-over-union~(mIoU) as the metric.

\myPara{Foreground Segmentation.} We evaluate our model on the Pascal-5i dataset~\cite{shaban2017one}, including 4 different test splits. Following the protocol in~\citep{bar2022visual}, we extract binary masks from our predictions and report the mIoU. 

\myPara{Deraining.} We randomly sample 1,000 images from ImageNet-1K validation set~\cite{deng2009imagenet} and apply the raining filter on them. We adopt the Peak Signal-to-Noise Ratio (PSNR) and Structural Similarity Index Measure (SSIM) as metrics.

\myPara{De-motion Blur.} We randomly sample 1,000 images from the ImageNet-1K validation set~\cite{deng2009imagenet} and apply motion blur on these images. We adopt the PSNR and SSIM as metrics.

\subsection{Architecture Details of LaVin-DiT}\label{appendix:arch_details}
Here we detail the architecture of the LaVin-DiT models. Table~\ref{tab:model_detail} outlines the configurations for three parameter scales: 0.1B, 1.0B, and 3.4B. Each configuration is characterized by key architectural hyperparameters, including the number of latent channels, patch size, hidden channels, and the number of layers. Additionally, the configurations specify the number of attention heads, key-value groups, drop path rates, and unconditional ratios. To further enhance model training, we incorporate advanced techniques such as gradient clipping and the Exponential Moving Average (EMA). All models utilize 3D-RoPE  to ensure consistent spatial and temporal encoding across scales. For large models, we employ sandwich normalization to improve training stability.

\begin{figure*}[t]
    \centering
    \includegraphics[width=1.\linewidth]{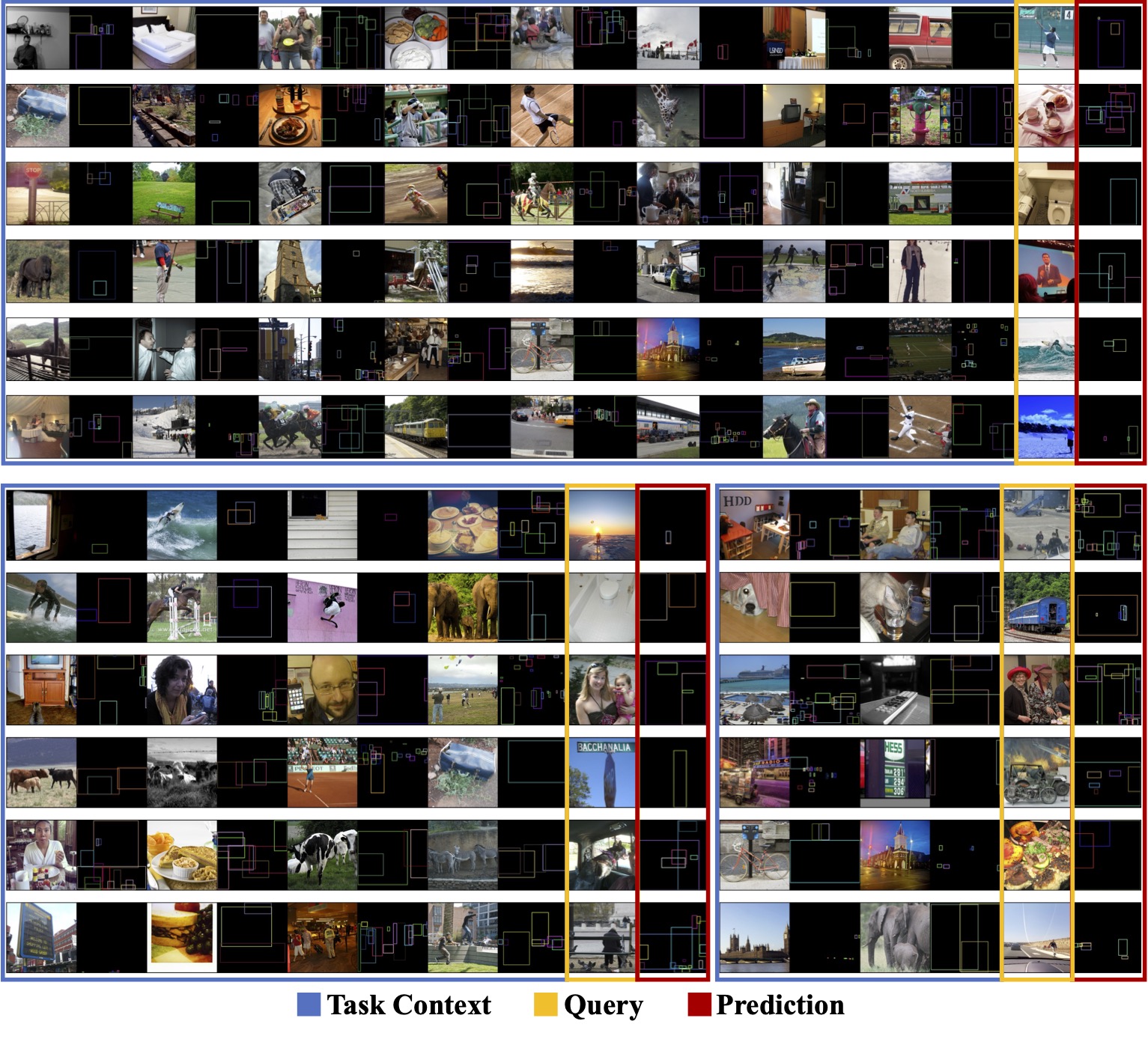}
    \caption{\textbf{Qualitative results on object detection.} Each row contains a sequence of images interleaved with annotations, followed by a query. The last image is predicted by the model (marked in red). \textit{Best viewed in color.}}
    \vspace{-0.1cm}
    \label{fig:vis_det}
\end{figure*}

\section{Supplementary Qualitative Results}\label{appendix:more_exp_results}
\begin{figure*}[t]
    \centering
    \includegraphics[width=1.\linewidth]{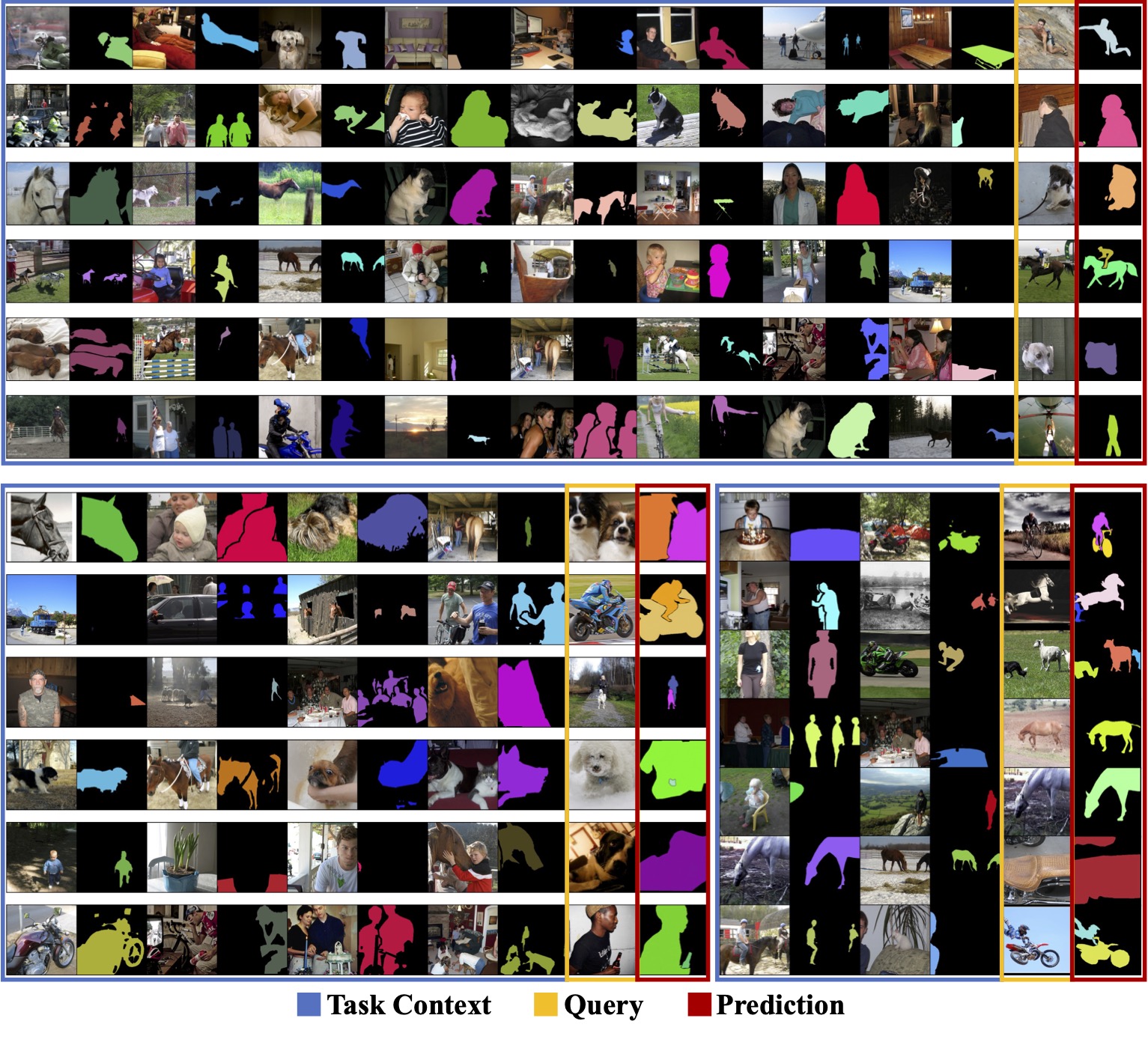}
    \caption{\textbf{Qualitative results on foreground segmentation.} Each row contains a sequence of images interleaved with annotations, followed by a query. The last image is predicted by the model (marked in red). \textit{Best viewed in color.}}
    \vspace{-0.385cm}
    \label{fig:vis_fseg}
\end{figure*}

\begin{figure*}[t]
    \centering
    \includegraphics[width=1.\linewidth]{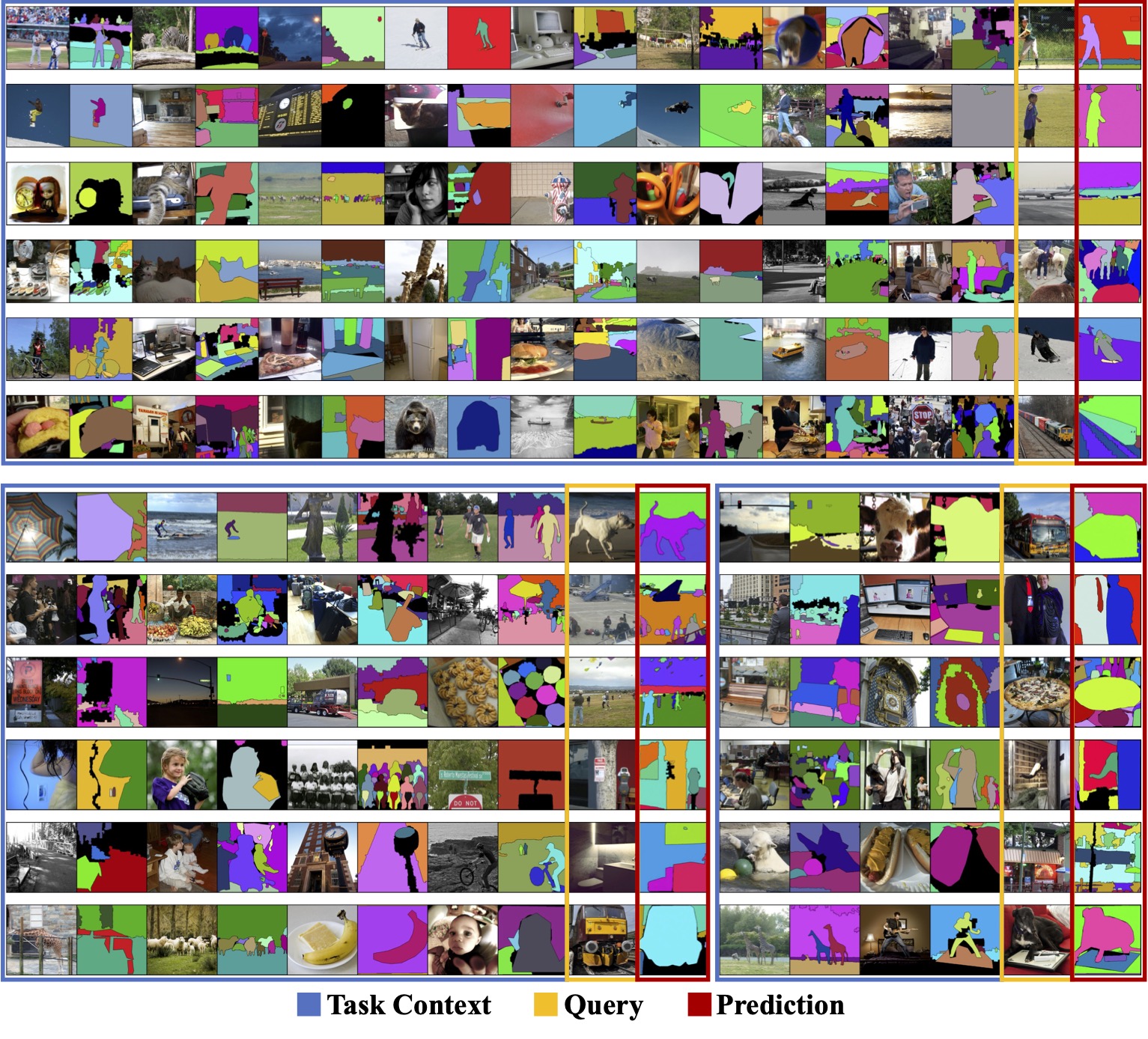}
    \caption{\textbf{Qualitative results on panoptic segmentation.} Each row contains a sequence of images interleaved with annotations, followed by a query. The last image is predicted by the model (marked in red). \textit{Best viewed in color.}}
    \vspace{-0.385cm}
    \label{fig:vis_pseg}
\end{figure*}

\begin{figure*}[t]
    \centering
    \includegraphics[width=1.\linewidth]{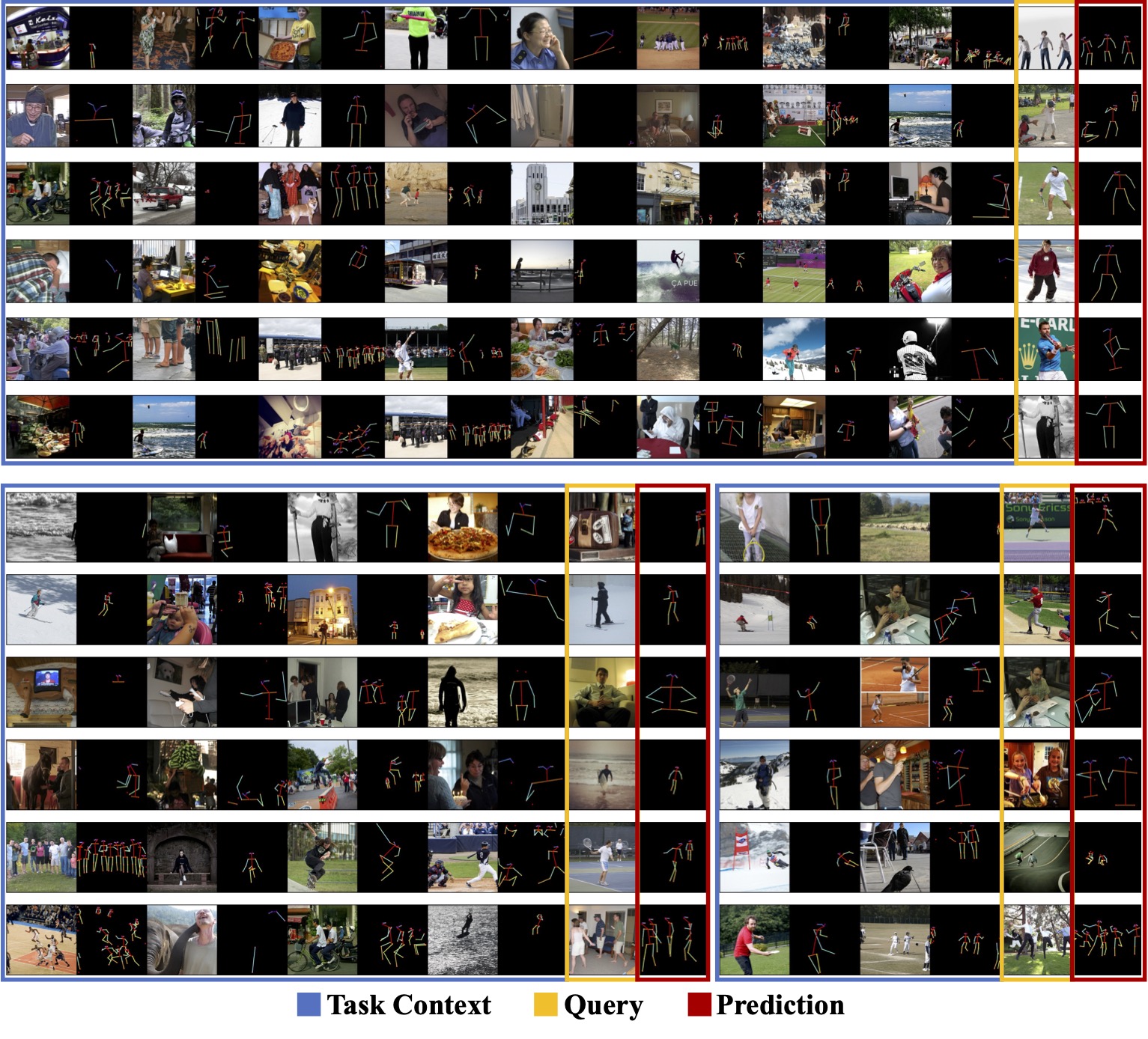}
    \caption{\textbf{Qualitative results on pose estimation.} Each row contains a sequence of images interleaved with annotations, followed by a query. The last image is predicted by the model (marked in red). \textit{Best viewed in color.}}
    \vspace{-0.385cm}
    \label{fig:vis_pose}
\end{figure*}

\begin{figure*}[t]
    \centering
    \includegraphics[width=1.\linewidth]{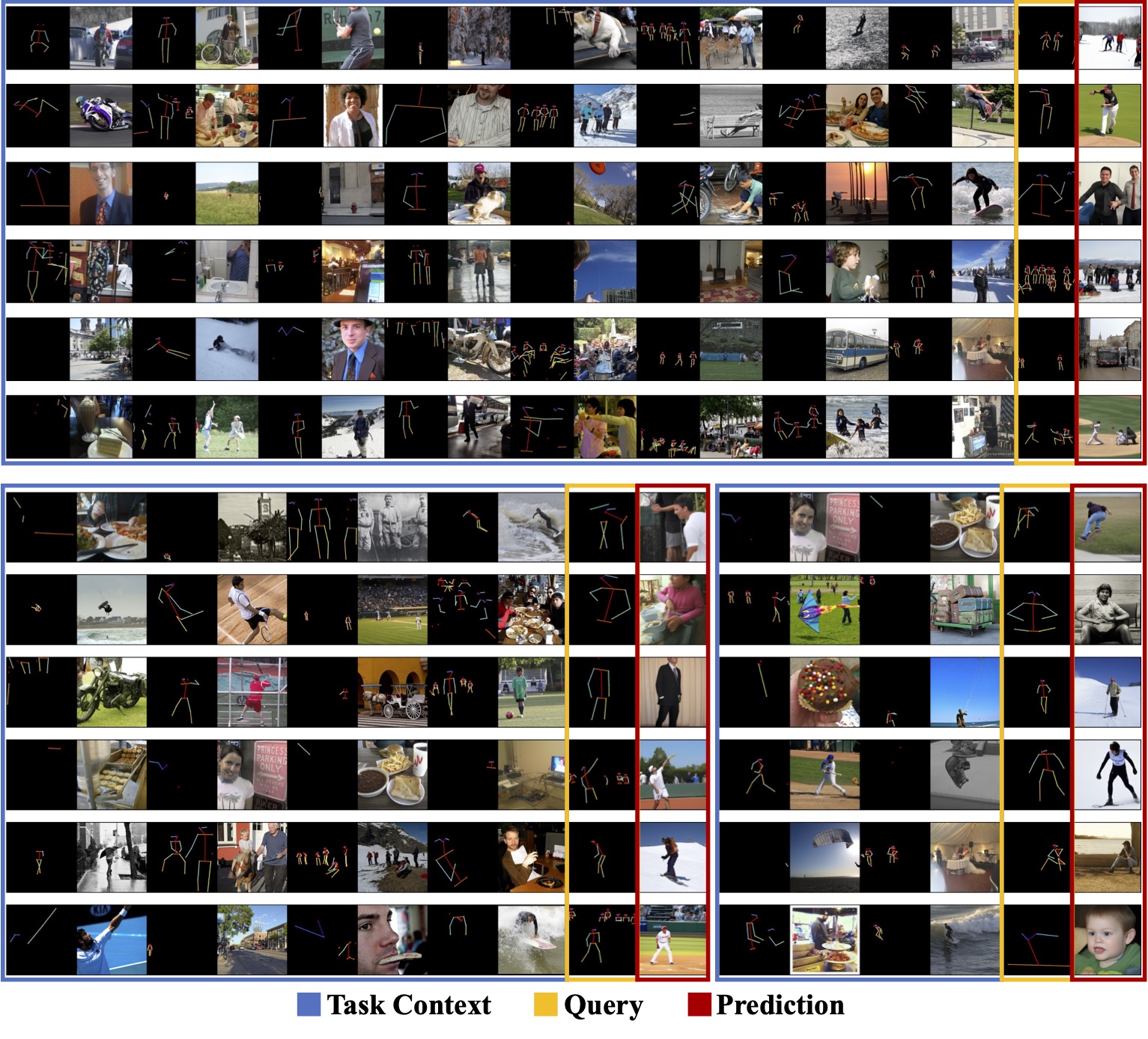}
    \caption{\textbf{Qualitative results on pose-to-image generation.} Each row contains a sequence of images interleaved with annotations, followed by a query. The last image is predicted by the model (marked in red). \textit{Best viewed in color.}}
    \vspace{-0.385cm}
    \label{fig:vis_reverse_pose}
\end{figure*}

\begin{figure*}[t]
    \centering
    \includegraphics[width=1.\linewidth]{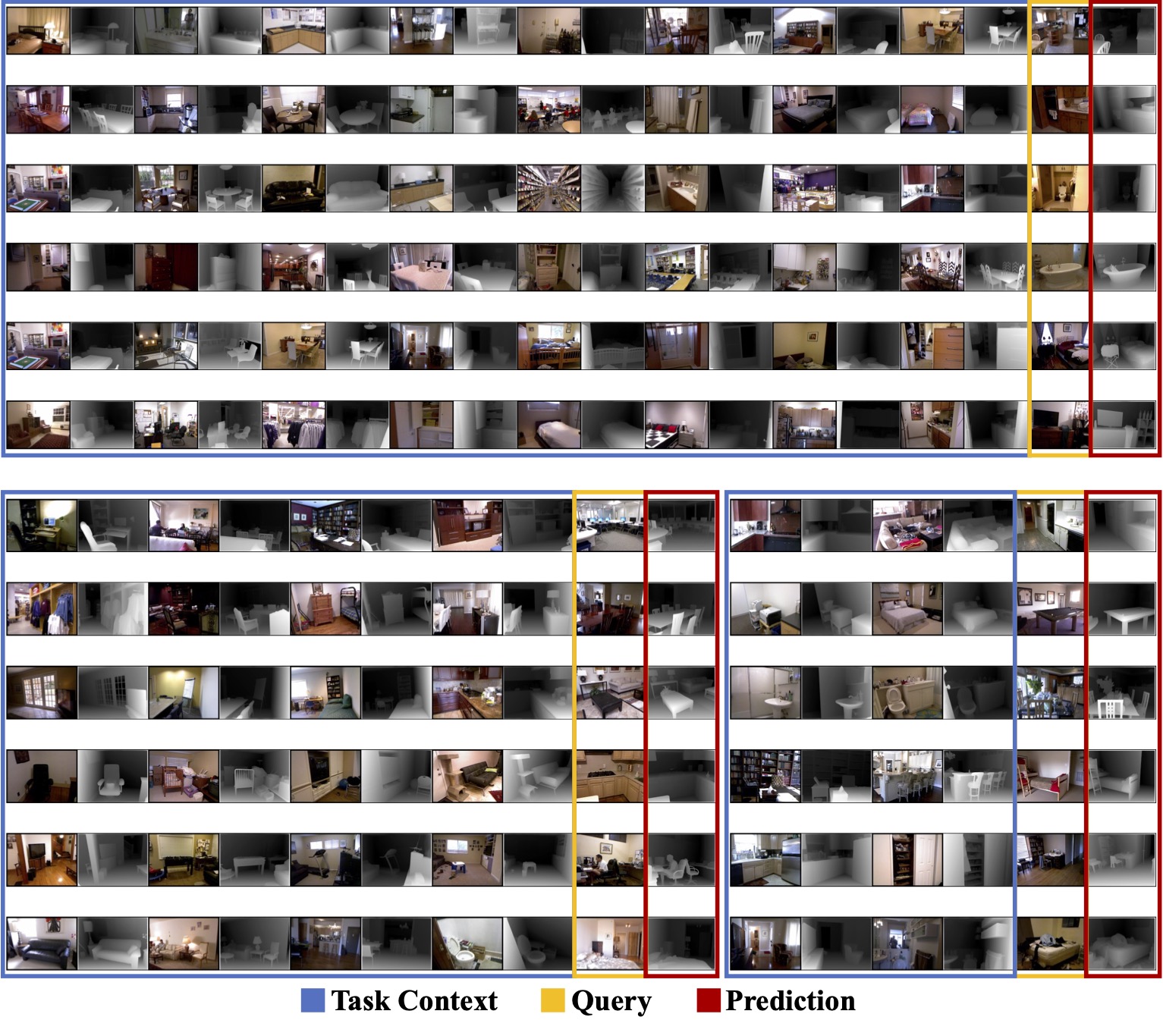}
    \caption{\textbf{Qualitative results on depth estimation.} Each row contains a sequence of images interleaved with annotations, followed by a query. The last image is predicted by the model (marked in red). \textit{Best viewed in color.}}
    \vspace{-0.385cm}
    \label{fig:vis_depth}
\end{figure*}

\begin{figure*}[t]
    \centering
    \includegraphics[width=1.\linewidth]{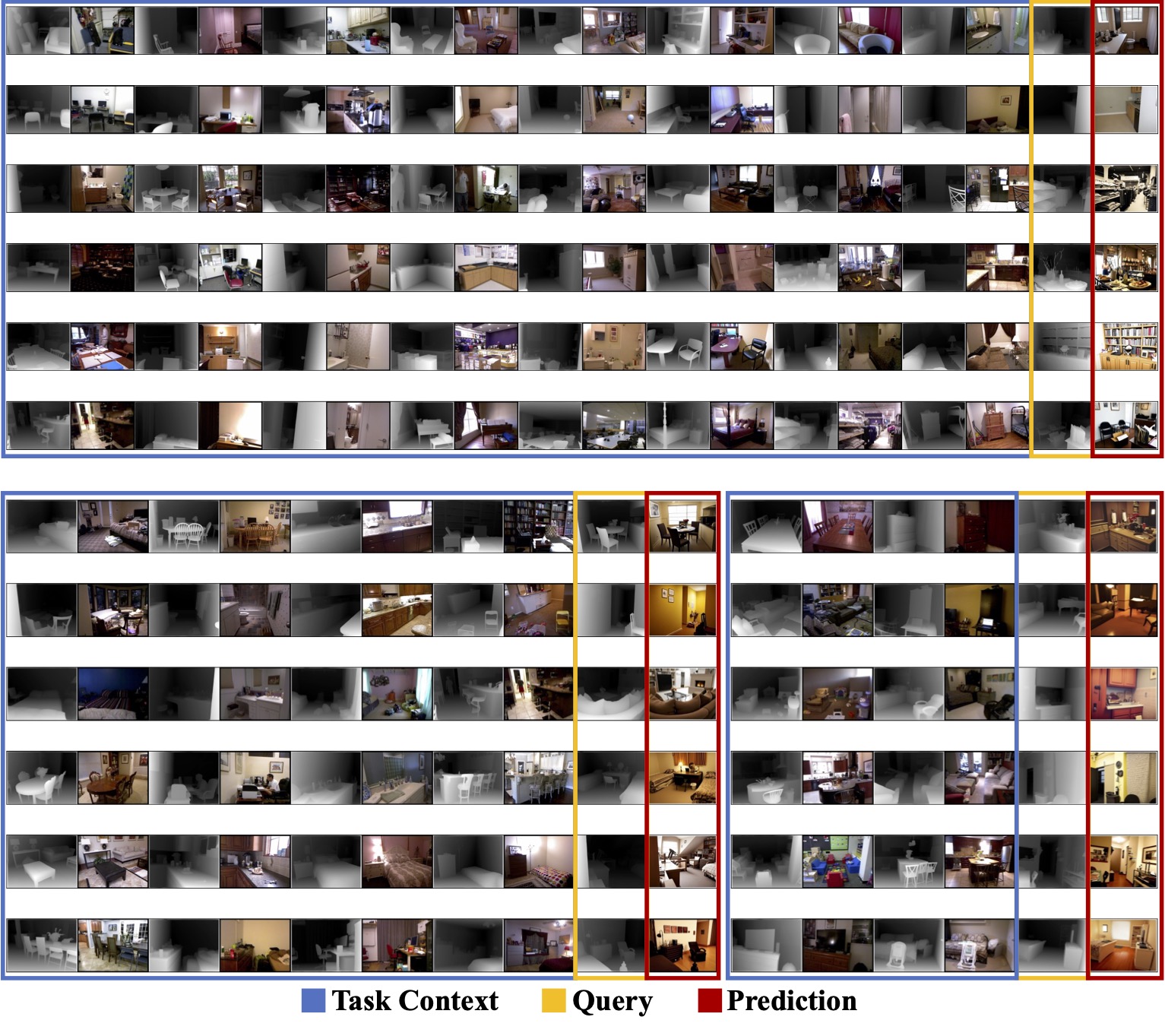}
    \caption{\textbf{Qualitative results on depth-to-image generation.} Each row contains a sequence of images interleaved with annotations, followed by a query. The last image is predicted by the model (marked in red). \textit{Best viewed in color.}}
    \vspace{-0.385cm}
    \label{fig:vis_reverse_depth}
\end{figure*}

\begin{figure*}[t]
    \centering
    \includegraphics[width=1.\linewidth]{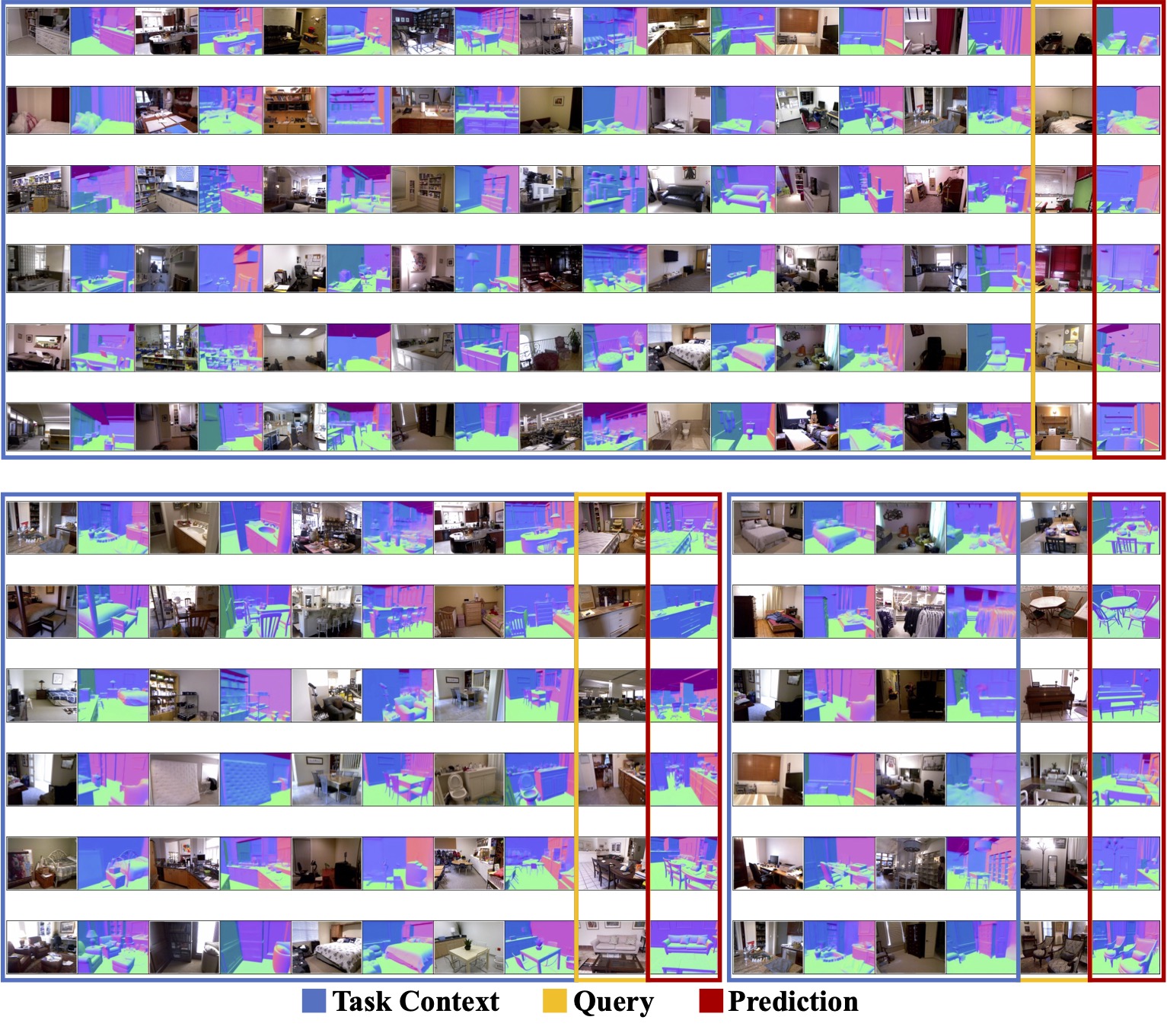}
    \caption{\textbf{Qualitative results on surface normal estimation.} Each row contains a sequence of images interleaved with annotations, followed by a query. The last image is predicted by the model (marked in red). \textit{Best viewed in color.}}
    \vspace{-0.385cm}
    \label{fig:vis_normal}
\end{figure*}

\begin{figure*}[t]
    \centering
    \includegraphics[width=1.\linewidth]{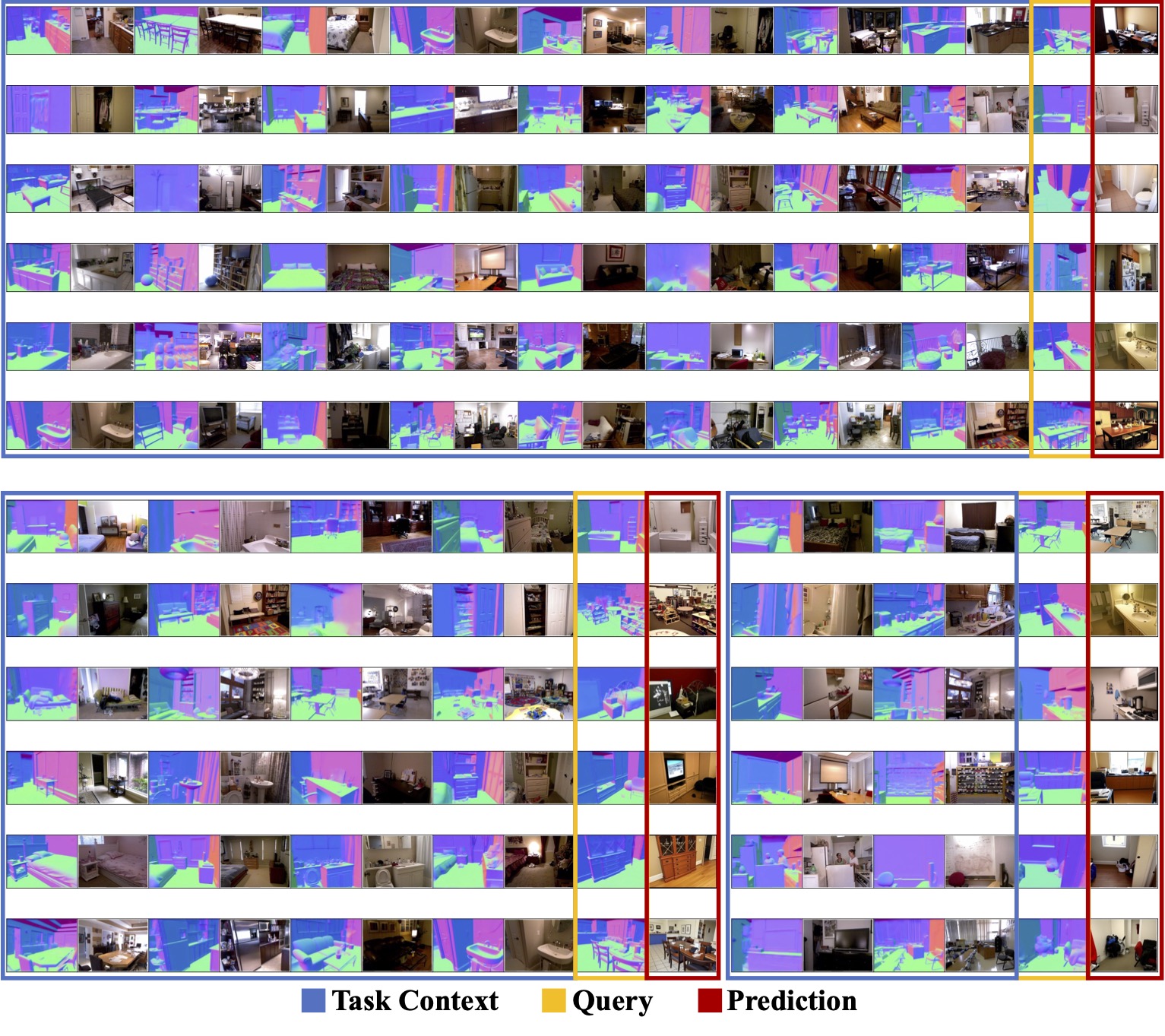}
    \caption{\textbf{Qualitative results on normal-to-image generation.} Each row contains a sequence of images interleaved with annotations, followed by a query. The last image is predicted by the model (marked in red). \textit{Best viewed in color.}}
    \vspace{-0.385cm}
    \label{fig:vis_reverse_normal}
\end{figure*}

\begin{figure*}[t]
    \centering
    \includegraphics[width=1.\linewidth]{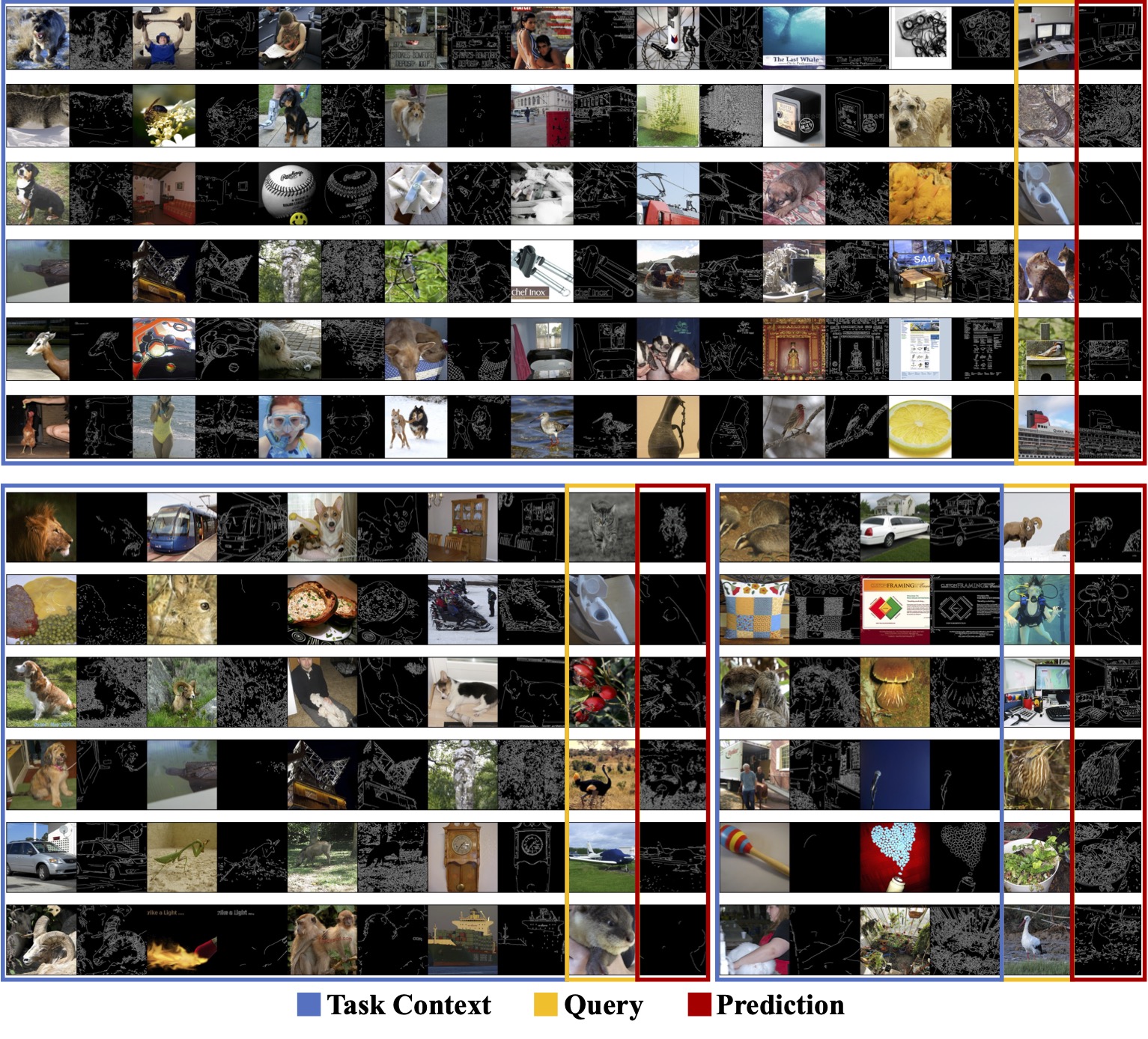}
    \caption{\textbf{Qualitative results on edge detection.} Each row contains a sequence of images interleaved with annotations, followed by a query. The last image is predicted by the model (marked in red). \textit{Best viewed in color.}}
    \vspace{-0.385cm}
    \label{fig:vis_edge}
\end{figure*}

\begin{figure*}[t]
    \centering
    \includegraphics[width=1.\linewidth]{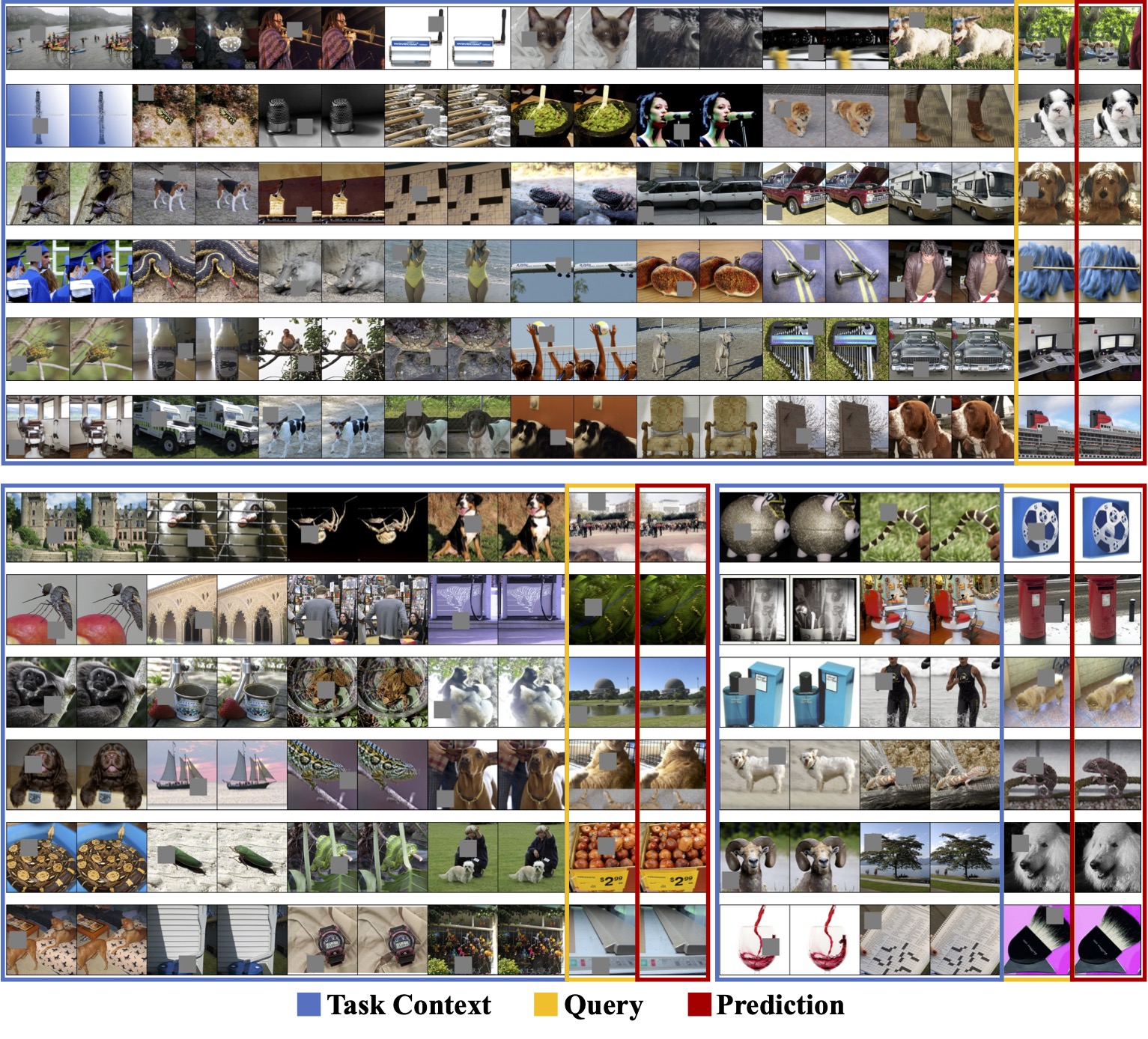}
    \caption{\textbf{Qualitative results on inpainting.} Each row contains a sequence of images interleaved with annotations, followed by a query. The last image is predicted by the model (marked in red). \textit{Best viewed in color.}}
    \vspace{-0.385cm}
    \label{fig:vis_inpaint}
\end{figure*}

\begin{figure*}[t]
    \centering
    \includegraphics[width=1.\linewidth]{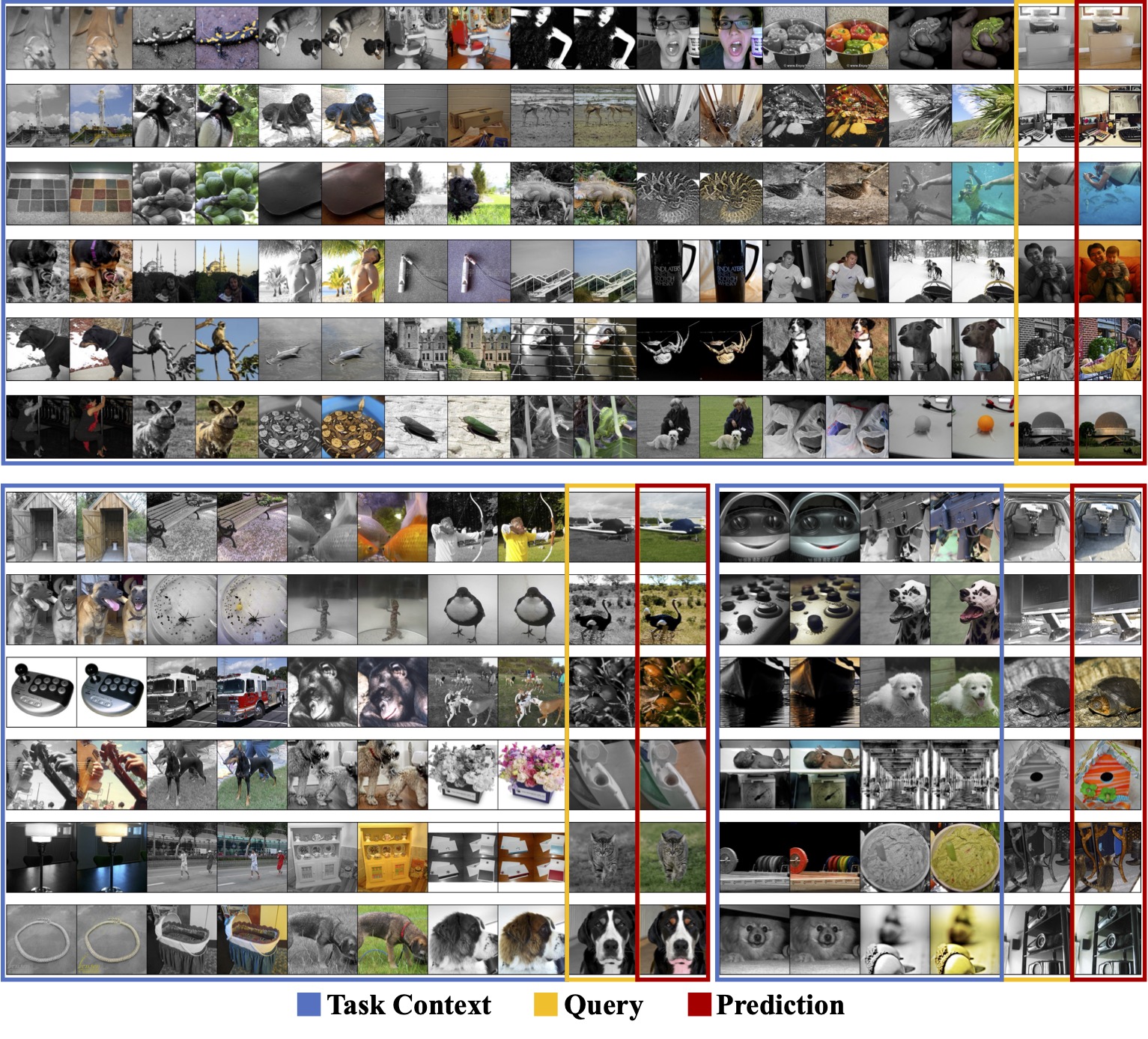}
    \caption{\textbf{Qualitative results on image colorization.} Each row contains a sequence of images interleaved with annotations, followed by a query. The last image is predicted by the model (marked in red). \textit{Best viewed in color.}}
    \vspace{-0.385cm}
    \label{fig:vis_color}
\end{figure*}

\begin{figure*}[t]
    \centering
    \includegraphics[width=1.\linewidth]{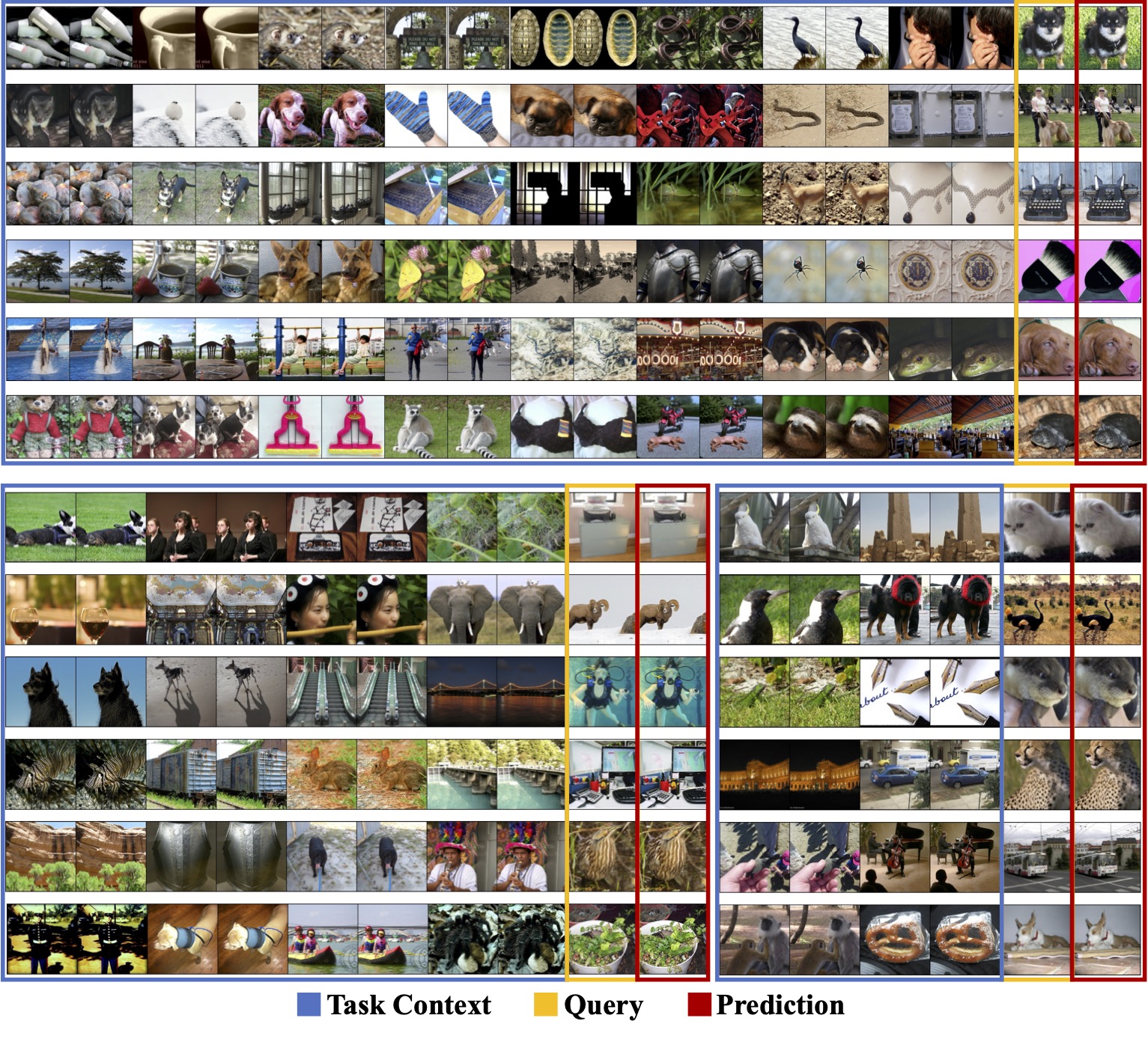}
    \caption{\textbf{Qualitative results on de-glass blur.} Each row contains a sequence of images interleaved with annotations, followed by a query. The last image is predicted by the model (marked in red). \textit{Best viewed in color.}}
    \vspace{-0.385cm}
    \label{fig:vis_degblur}
\end{figure*}

\begin{figure*}[t]
    \centering
    \includegraphics[width=1.\linewidth]{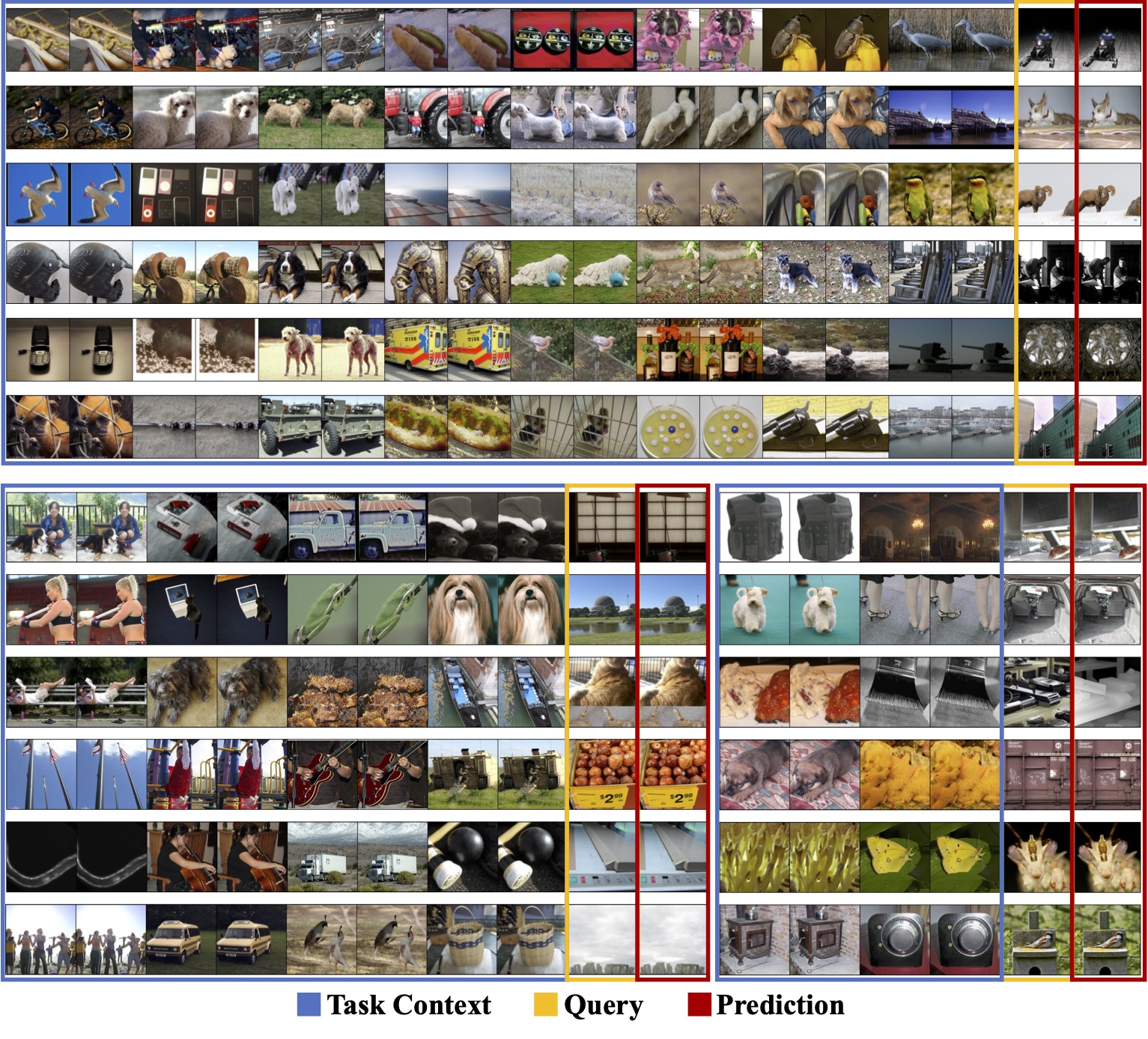}
    \caption{\textbf{Qualitative results on de-motion blur.} Each row contains a sequence of images interleaved with annotations, followed by a query. The last image is predicted by the model (marked in red). \textit{Best viewed in color.}}
    \vspace{-0.385cm}
    \label{fig:vis_demblur}
\end{figure*}

\begin{figure*}[t]
    \centering
    \includegraphics[width=1.\linewidth]{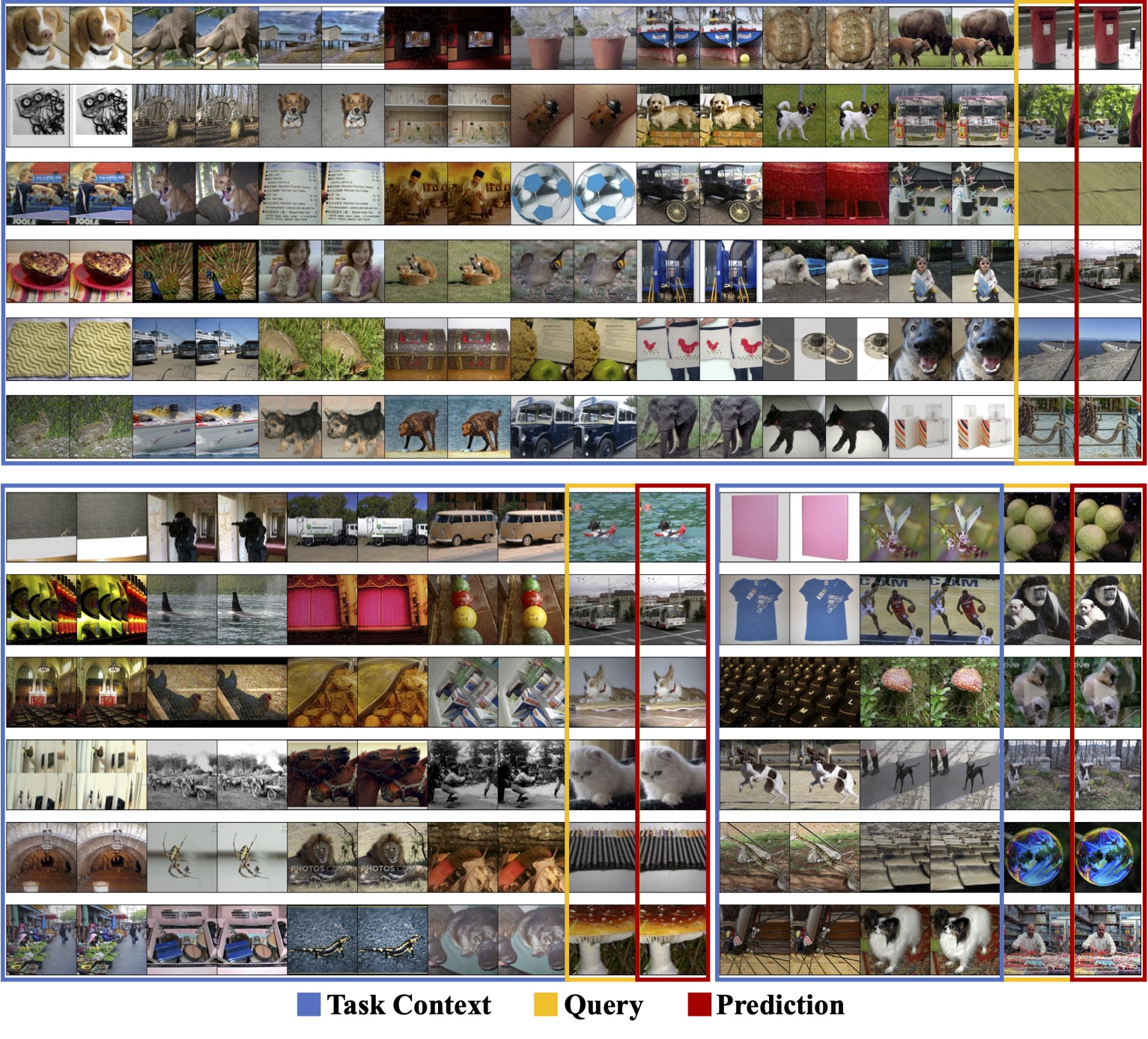}
    \caption{\textbf{Qualitative results on de-raining.} Each row contains a sequence of images interleaved with annotations, followed by a query. The last image is predicted by the model (marked in red). \textit{Best viewed in color.}}
    \vspace{-0.385cm}
    \label{fig:vis_derain}
\end{figure*}

\begin{figure*}[t]
    \centering
    \includegraphics[width=1.\linewidth]{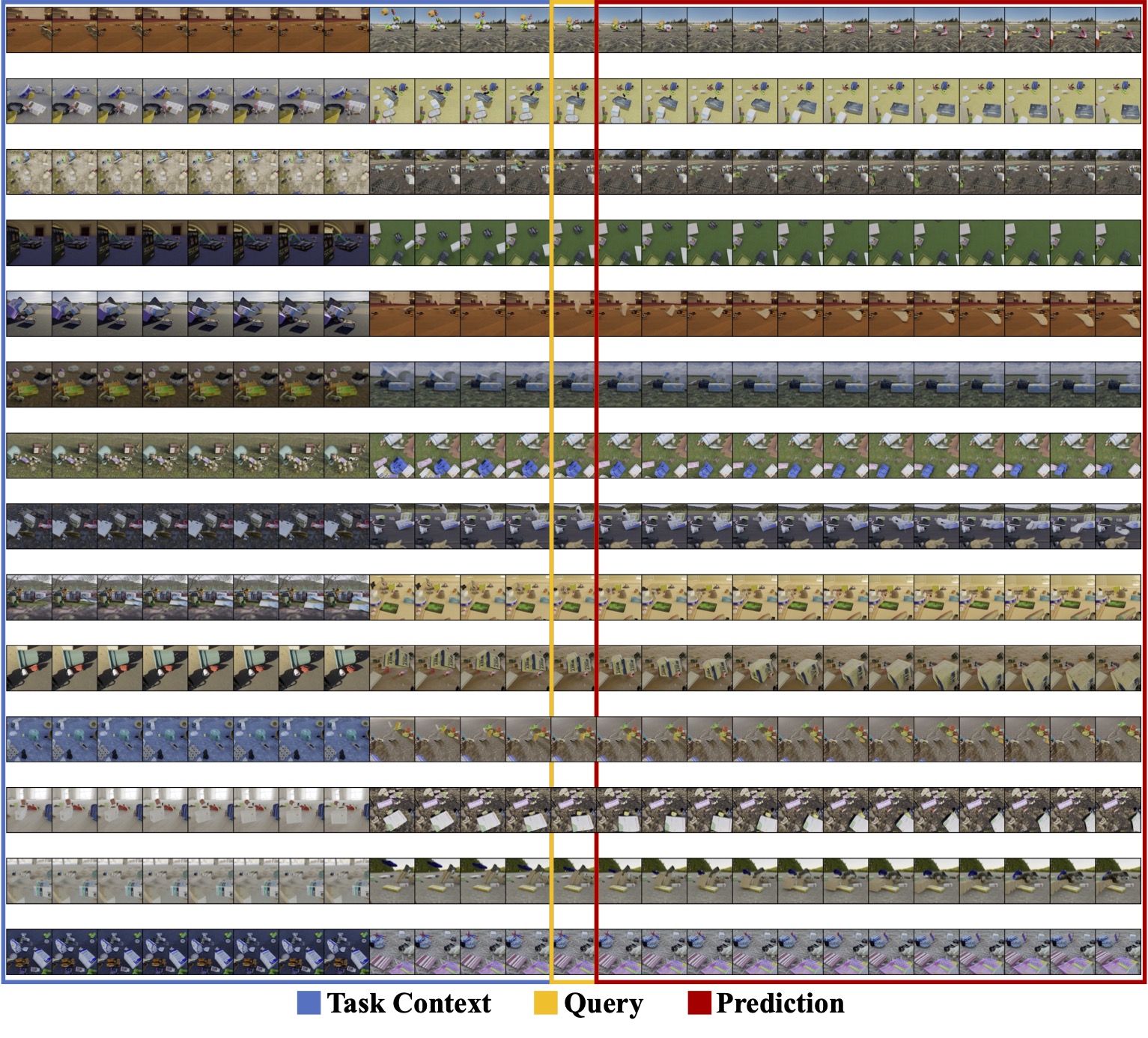}
    \caption{\textbf{Qualitative results on frame prediction.} Each row includes a video sequence with a series of target frames as task context (marked in blue), followed by a query frame (marked in yellow). A set of frames in the red box indicates the model’s predictions. Due to the length of the sequence, a portion of the task context is hidden. \textit{Best viewed in color.}}
    \vspace{-0.385cm}
    \label{fig:vis_vidpred}
\end{figure*}

\begin{figure*}[t]
    \centering
    \includegraphics[width=1.\linewidth]{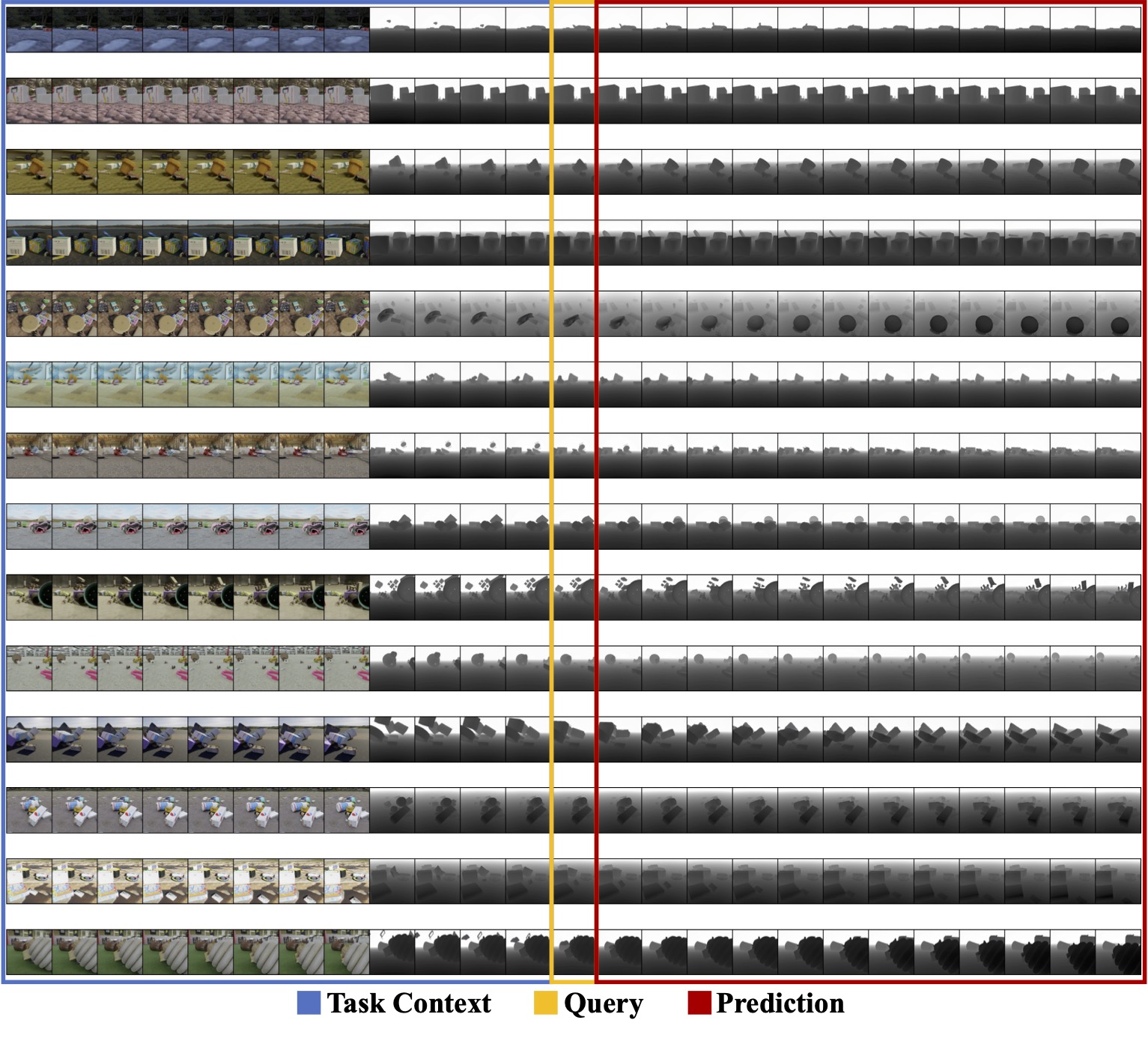}
    \caption{\textbf{Qualitative results on video depth estimation.} Each row includes a video sequence with a series of target frames as task context (marked in blue), followed by a query frame (marked in yellow). A set of frames in the red box indicates the model’s predictions. Due to the length of the sequence, a portion of the task context is hidden. \textit{Best viewed in color.}}
    \vspace{-0.385cm}
    \label{fig:vis_viddepth}
\end{figure*}

\begin{figure*}[t]
    \centering
    \includegraphics[width=1.\linewidth]{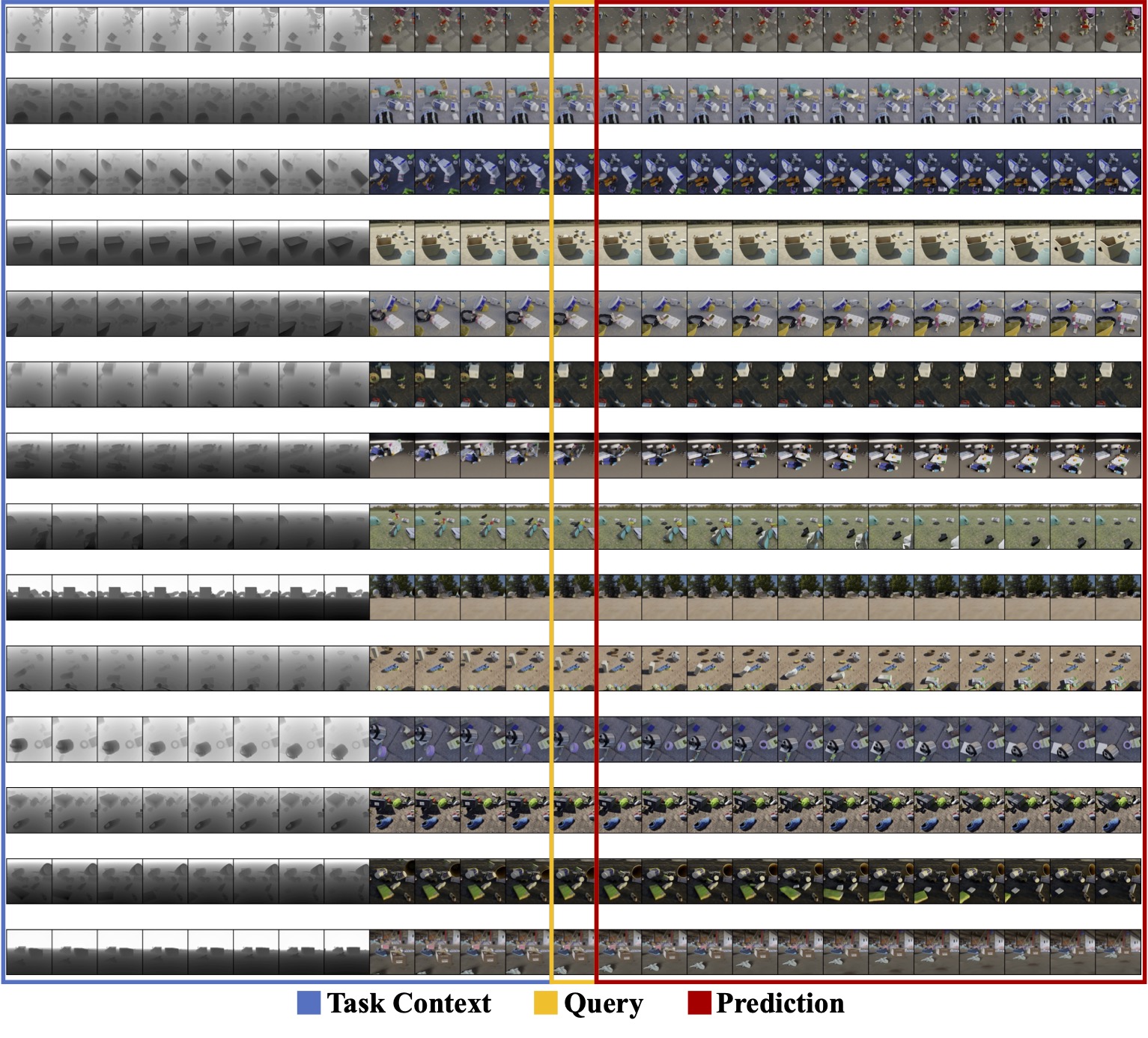}
    \caption{\textbf{Qualitative results on depth-to-video generation.} Each row includes a video sequence with a series of target frames as task context (marked in blue), followed by a query frame (marked in yellow). A set of frames in the red box indicates the model’s predictions. Due to the length of the sequence, a portion of the task context is hidden. \textit{Best viewed in color.}}
    \vspace{-0.385cm}
    \label{fig:vis_vid-reversedepth}
\end{figure*}

\begin{figure*}[t]
    \centering
    \includegraphics[width=1.\linewidth]{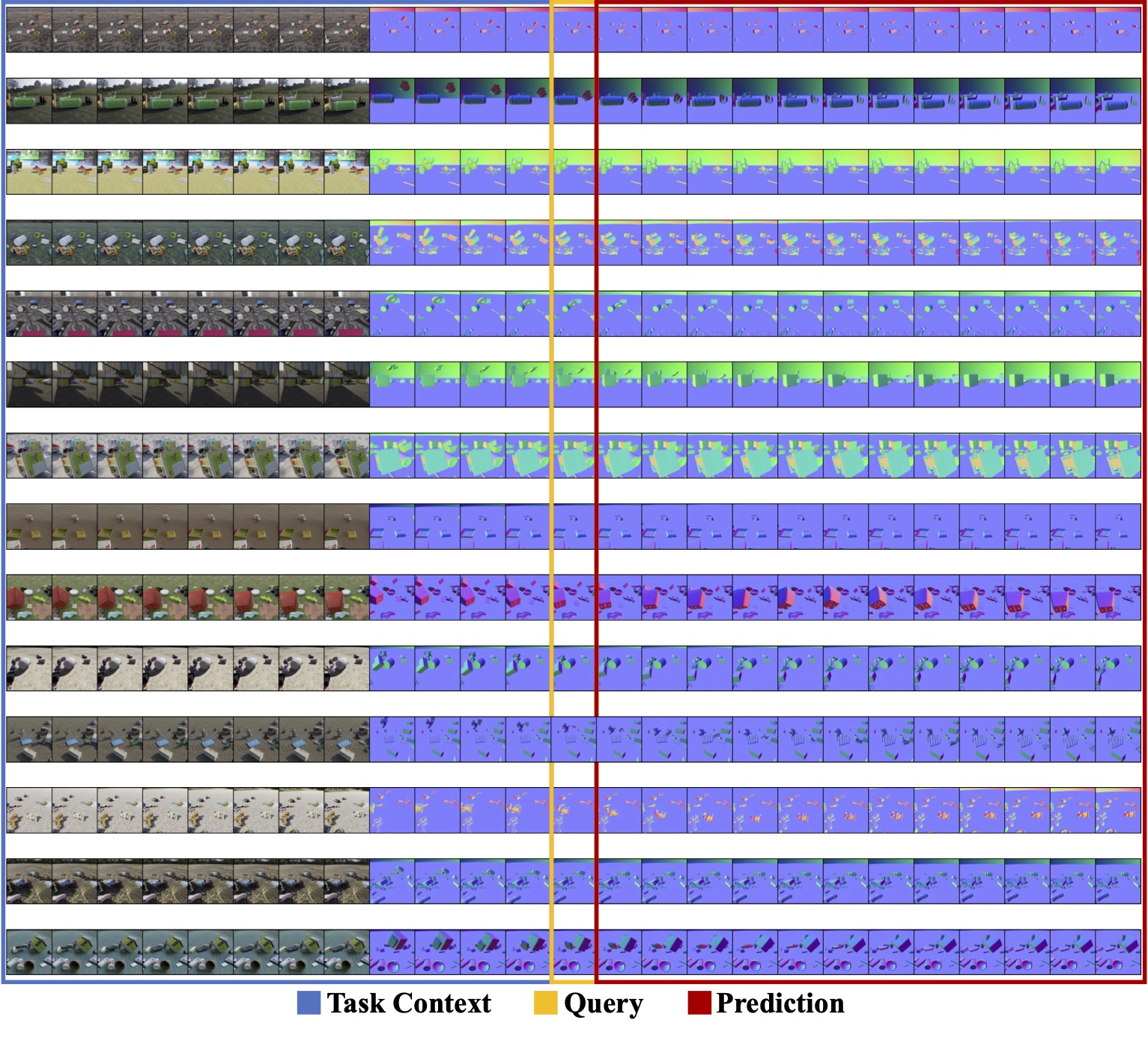}
    \caption{\textbf{Qualitative results on video surface normal estimation.} Each row includes a video sequence with a series of target frames as task context (marked in blue), followed by a query frame (marked in yellow). A set of frames in the red box indicates the model’s predictions. Due to the length of the sequence, a portion of the task context is hidden. \textit{Best viewed in color.}}
    \vspace{-0.385cm}
    \label{fig:vis_vidnormal}
\end{figure*}

\begin{figure*}[t]
    \centering
    \includegraphics[width=1.\linewidth]{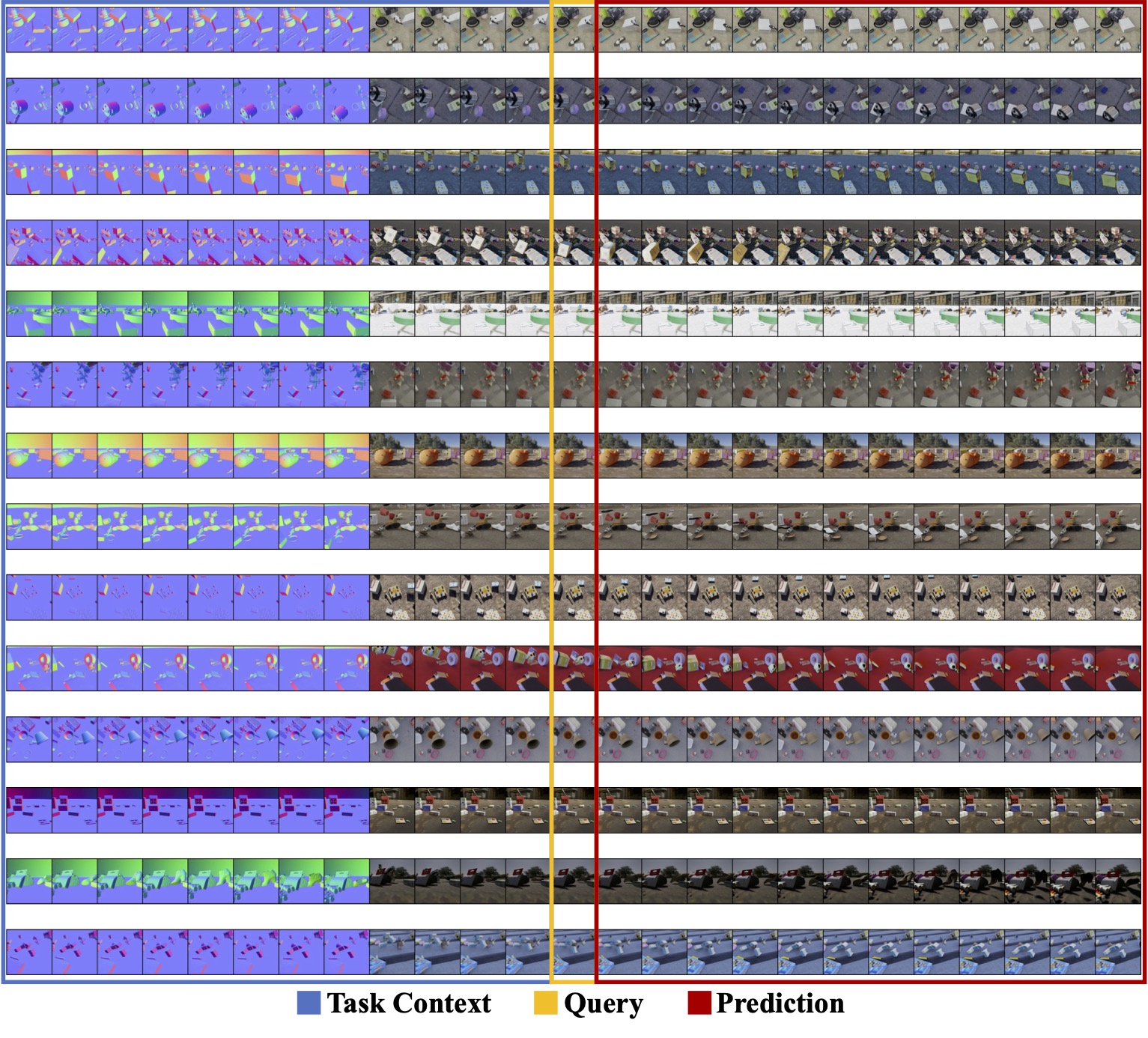}
    \caption{\textbf{Qualitative results on normal-to-video generation.} Each row includes a video sequence with a series of target frames as task context (marked in blue), followed by a query frame (marked in yellow). A set of frames in the red box indicates the model’s predictions. Due to the length of the sequence, a portion of the task context is hidden. \textit{Best viewed in color.}}
    \vspace{-0.385cm}
    \label{fig:vis_vid-reversenormal}
\end{figure*}

\begin{figure*}[t]
    \centering
    \includegraphics[width=1.\linewidth]{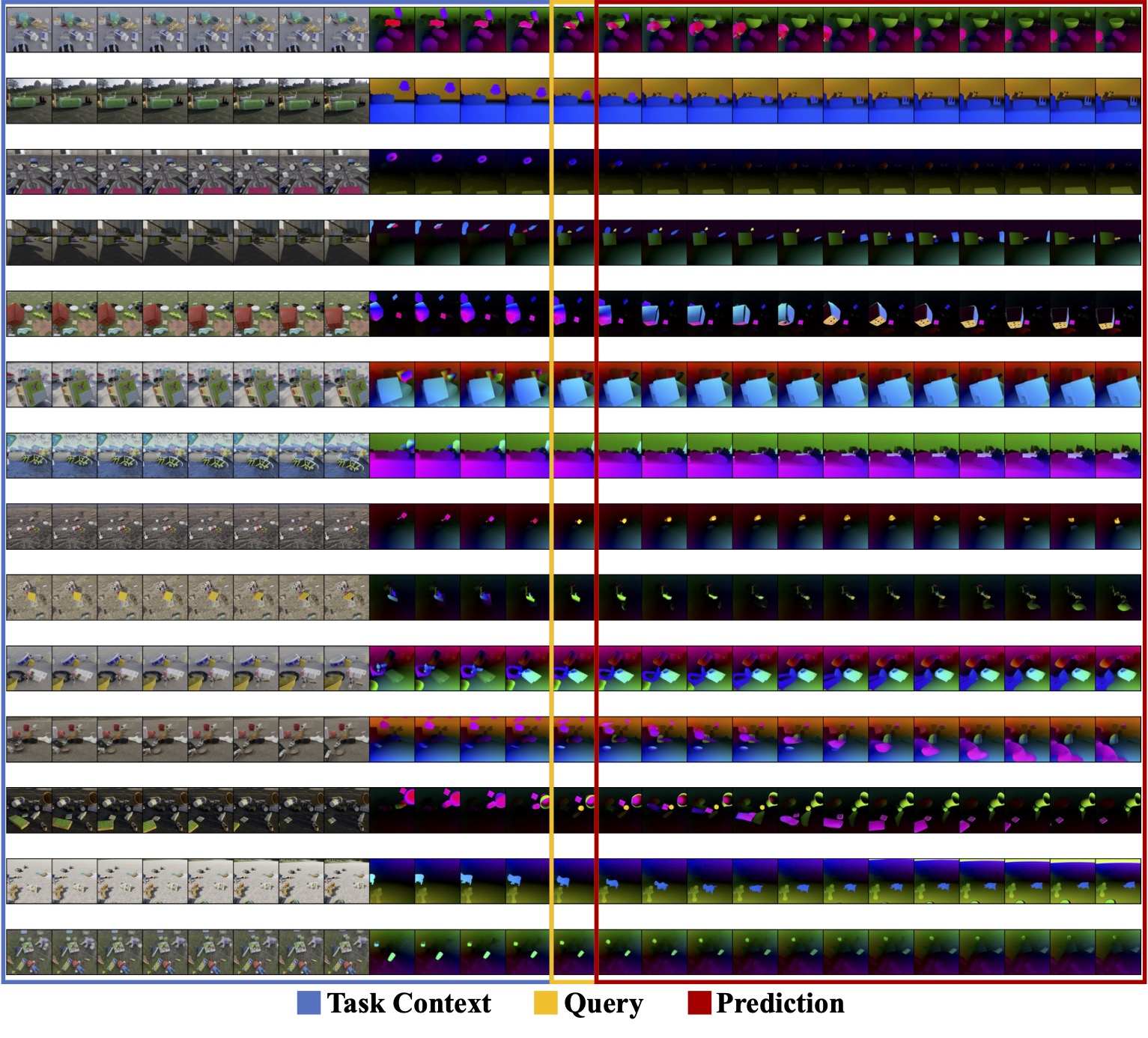}
    \caption{\textbf{Qualitative results on optical flow estimation.} Each row includes a video sequence with a series of target frames as task context (marked in blue), followed by a query frame (marked in yellow). A set of frames in the red box indicates the model’s predictions. Due to the length of the sequence, a portion of the task context is hidden. \textit{Best viewed in color.}}
    \vspace{-0.385cm}
    \label{fig:vis_vidflow}
\end{figure*}

\begin{figure*}[t]
    \centering
    \includegraphics[width=1.\linewidth]{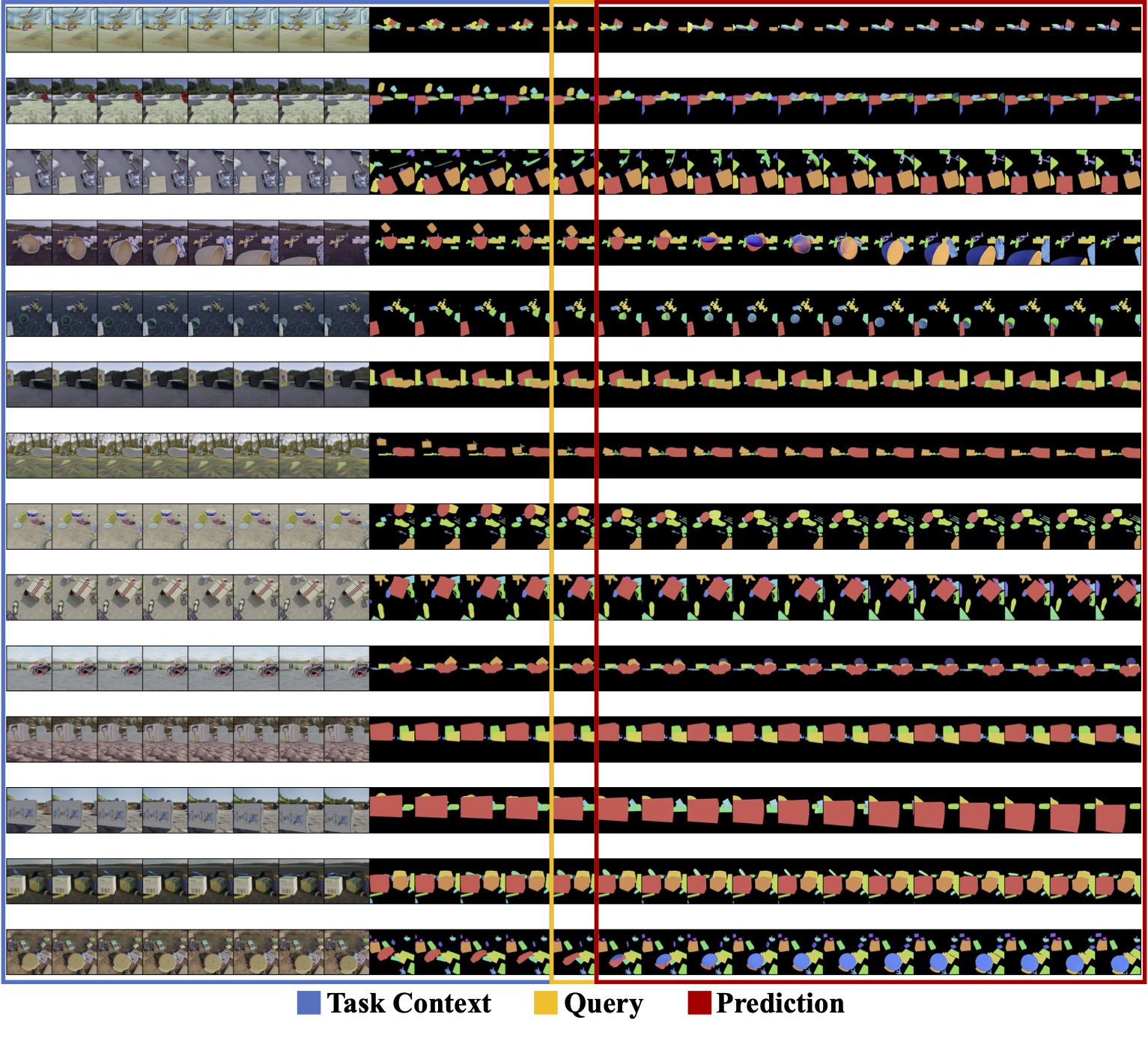}
    \caption{\textbf{Qualitative results on video instance segmentation.} Each row includes a video sequence with a series of target frames as task context (marked in blue), followed by a query frame (marked in yellow). A set of frames in the red box indicates the model’s predictions. Due to the length of the sequence, a portion of the task context is hidden. \textit{Best viewed in color.}}
    \vspace{-0.385cm}
    \label{fig:vis_vidsegmentation}
\end{figure*}

We show more visualization results for each task, including object detection (Figure~\ref{fig:vis_det}), foreground segmentation (Figure~\ref{fig:vis_fseg}), panoptic segmentation (Figure~\ref{fig:vis_pseg}), pose estimation (Figure~\ref{fig:vis_pose}), pose-to-image generation (Figure~\ref{fig:vis_reverse_pose}), depth estimation (Figure~\ref{fig:vis_depth}), depth-to-image generation (Figure~\ref{fig:vis_reverse_depth}), surface normal estimation (Figure~\ref{fig:vis_normal}), normal-to-image generation (Figure~\ref{fig:vis_reverse_normal}), edge detection (Figure~\ref{fig:vis_edge}), inpainting (Figure~\ref{fig:vis_inpaint}), colorization (Figure~\ref{fig:vis_color}), de-glass blur (Figure~\ref{fig:vis_degblur}), de-motion blur (Figure~\ref{fig:vis_demblur}), de-raining (Figure~\ref{fig:vis_derain}), frame prediction (Figure~\ref{fig:vis_vidpred}), video depth estimation (Figure~\ref{fig:vis_viddepth}), depth-to-video generation (Figure~\ref{fig:vis_vid-reversedepth}), video surface normal estimation (Figure~\ref{fig:vis_vidnormal}), normal-to-video generation (Figure~\ref{fig:vis_vid-reversenormal}), video optical flow estimation (Figure~\ref{fig:vis_vidflow}), and video instance segmentation (Figure~\ref{fig:vis_vidsegmentation}).


\section{Potential Applications}\label{appendix:applications}
\begin{figure*}[t]
    \centering
    \includegraphics[width=1.\linewidth]{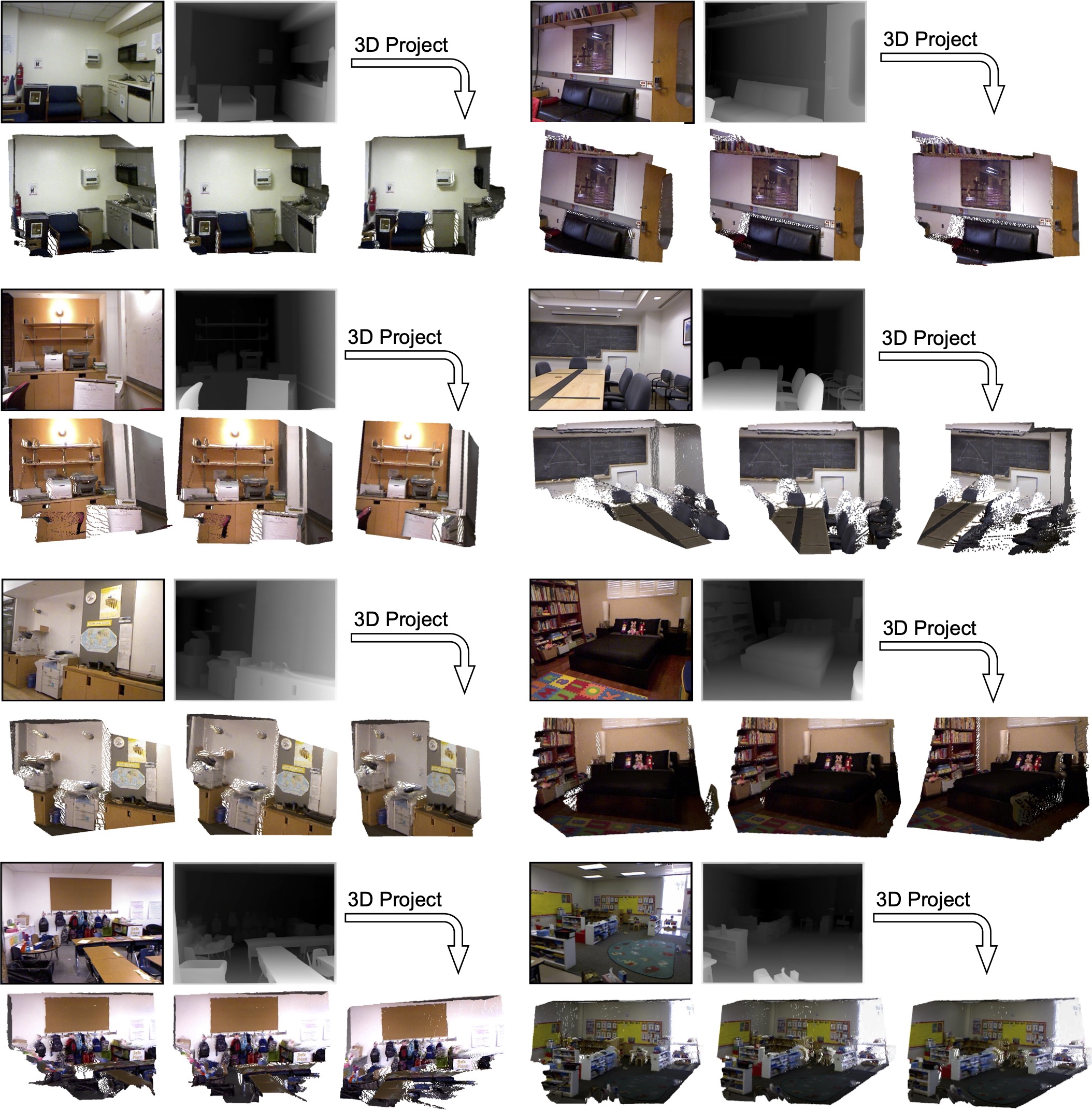}
    \caption{\textbf{Potential application of single-view scene reconstruction.} Given an RGB image and predicted depth map, we lift this image into a 3D space. We illustrate three views of this scene. \textit{Best viewed in color.}}
    \vspace{-0.385cm}
    \label{fig:vis_3drecon}
\end{figure*}

LaVin-DiT opens transformative possibilities for tackling open-world computer vision challenges by unifying diverse vision tasks within a single generative framework. For instance, it can seamlessly generalize across tasks such as text-to-image generation, text-to-video generation, video understanding, 3D reconstruction (Figure~\ref{fig:vis_3drecon}), and 2D/3D visual editing without supervised fine-tuning. By leveraging its spatial-temporal variational autoencoder and joint diffusion transformer, LaVin-DiT excels at capturing the complexity of high-dimensional visual data while maintaining task-specific alignment through in-context learning. This capability positions LaVin-DiT as a foundation model capable of addressing dynamic realistic vision problems, including autonomous driving perception, robotic scene understanding, and interactive AI systems in mixed-reality environments, significantly advancing the frontier of adaptable and scalable AI systems.
\clearpage

\end{document}